\pdfoutput=1

\documentclass[11pt]{article}

\usepackage[final]{acl}

\usepackage{times}
\usepackage{latexsym}

\usepackage[T1]{fontenc}
\usepackage[utf8]{inputenc}

\usepackage{microtype}

\usepackage{inconsolata}
\usepackage{graphicx}

\usepackage{hyperref}       
\usepackage{url}            
\usepackage{booktabs}       
\usepackage{amsfonts}       
\usepackage{nicefrac}       
\usepackage{microtype}      
\usepackage{xcolor}         
\usepackage{subfig}
\usepackage{pifont}
\usepackage{enumitem}
\usepackage{listings}
\usepackage{algorithm}
\usepackage{algpseudocode}
\usepackage{tcolorbox}
\usepackage{siunitx}  
\usepackage{threeparttable} 

\usepackage{colortbl}

\usepackage{graphicx}
\usepackage{multirow}
\usepackage{amsmath}
\usepackage{amssymb}
\usepackage{mathtools}
\usepackage{amsthm}
\usepackage{dsfont}
\usepackage{array}
\usepackage{makecell}

\title{Spectral Scaling Laws in Language Models: \\ {\em How Effectively Do Feed-Forward Networks Use Their Latent Space?}}

\author{%
  Nandan Kumar Jha \\
  New York University\\
  \texttt{nj2049@nyu.edu} \\
   \And
  Brandon Reagen \\
  New York University\\
  \texttt{bjr5@nyu.edu} \\
  \\
}

\begin{document}
\maketitle

\begin{abstract}

As Large Language Models (LLMs) scale, the question is not just how large they become, but {\em how much of their capacity is effectively utilized}. Existing scaling laws relate model size to loss, yet overlook how components exploit their latent space.  In this work, we focus on Feed-Forward Networks (FFNs) and recast width selection as a spectral utilization optimization problem. Using a lightweight diagnostic suite: Hard Rank (participation ratio), Soft Rank (Shannon Rank), Spectral Concentration, and the composite Spectral Utilization Index (SUI), we quantify how many latent directions are meaningfully activated across LLaMA, GPT-2, and nGPT families. 
Our {\em key finding} is an {\bf Asymmetric Spectral Scaling Law}: soft rank follows an almost perfect power law with FFN width, while hard rank grows only sublinearly, with high variance. This asymmetry suggests that widening FFNs mostly adds low-energy tail directions, while dominant-mode subspaces saturate early. Moreover, at larger widths, variance further collapses into a narrow subspace, leaving much of the latent space under-utilized. These results recast FFN width selection as a principled trade-off between tail capacity and dominant-mode capacity, offering concrete guidance for inference-efficient LLM design.
\end{abstract}

\section{Introduction}

As Large Language Models (LLMs) continue to grow in scale and complexity, a central blind spot remains: \textit{How effectively is their internal capacity utilized?} Existing empirical scaling laws \cite{kumar2025scaling,tao2024scaling,sardana2023beyond,kaplan2020scaling} relate model performance to factors such as width, depth, and data size, but they offer little insight into how different architectural components exploit, or potentially squander, the high-dimensional latent space. These laws treat models as black boxes, abstracting away the internal dynamics of transformer blocks and leaving open questions about representational usage.

\begin{figure} [t]
\centering
\includegraphics[width=.49\textwidth]{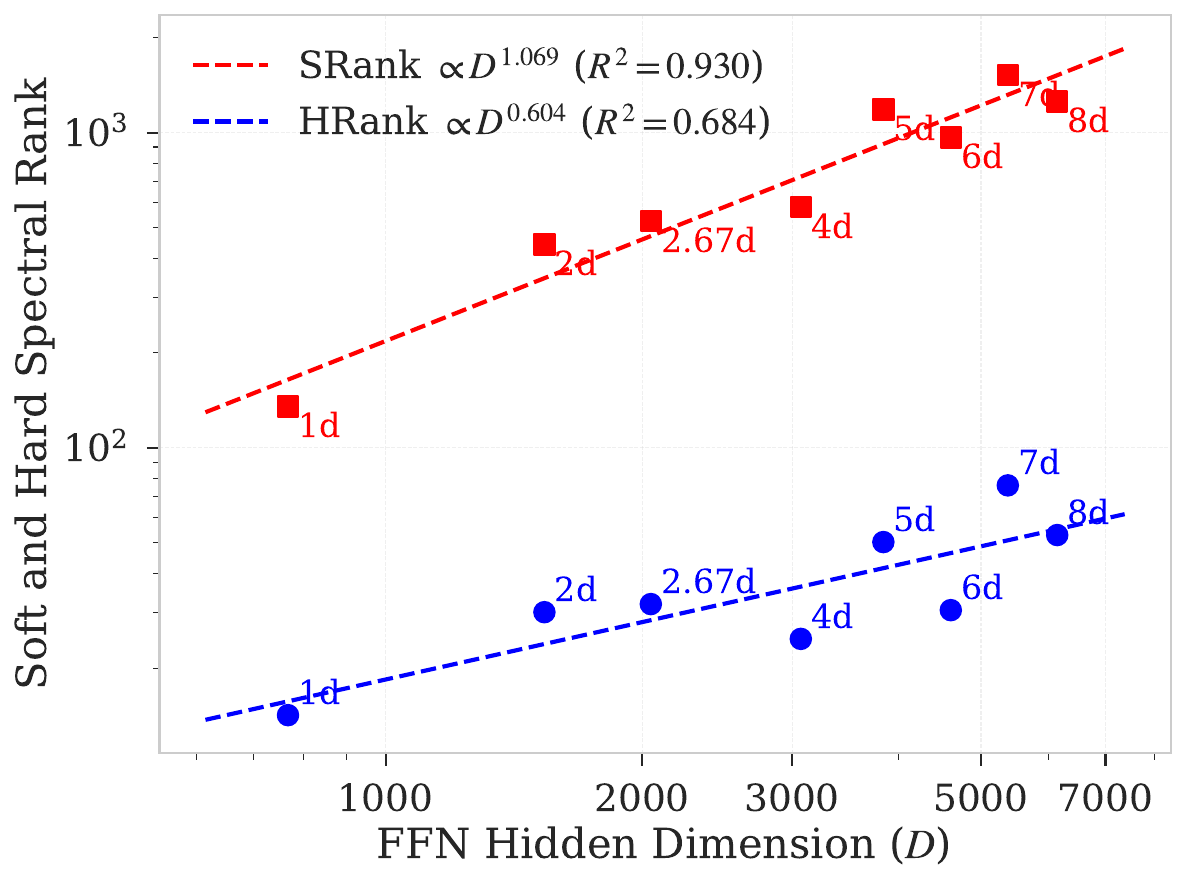}
\vspace{-2em}
\caption{Spectral rank vs. FFN hidden dimension in LLaMA-130M base model, with width sweep $D$ = $\alpha d$ (total parameters therefore differ across $\alpha$). Log-Log fits: Soft rank follows a linear power-law fit ($\beta$=1.06, $R^2$=0.93), while hard rank grows sublinearly ($\beta$=0.60, $R^2$=0.68), indicating width mainly adds low-energy tail directions rather than enlarging the high-energy dominant-mode subspace.}
\label{fig:ScalingLaws_Introduction}
\end{figure}

Among  transformer components, FFNs dominate the parameter budges as they can account for as much as 67\% of the total parameters in decoder-only models \cite{pires2023one,geva2020transformer}. Yet, FFN width is typically set by rules of thumb rather than design principles, e.g., 4$\times$ expansion in GPT-2 \cite{radford2019language} and 2.67$\times$ in LLaMA \cite{touvron2023llama}. Even in recent LLMs such as Qwen \cite{hui2024qwen2}, the FFN width  varies substantially across model sizes ($\approx$2.4-5.8$\times$)
underscoring the lack of theoretical grounding.

Despite their prevalence, we still lack a clear understanding of how FFN width affects effective capacity usage. This raises three questions: {\em Is increasing FFN width always beneficial for expressivity? How many latent directions are actually used in practice? Can we quantify representational efficiency beyond FLOPs and loss?}


We address these questions by reframing FFN width selection as a \textit{spectral utilization} problem. The {\bf intuition} is straightforward: if wider FFNs truly expand usable capacity, then their spectrum  should reflect growth in the effective dimensionality of the subspace the model exploits. To test this, we conduct a layer-wise spectral audit across GPT-2, LLaMA, and nGPT \cite{loshchilov2025ngpt} backbones, analyzing the eigenspectrum of post-activation covariance over training steps and layers.

We quantify utilization using four lightweight, differentiable metrics: Hard Rank (participation ratio) to capture the dimensionality of the high-energy, or the dominant, mode \cite{gao2017theory}; Soft Rank (Shannon Rank) to quantify uniformity across all directions \cite{de2016spectral}; Spectral Concentration (eigenvalue early enrichment) to quantify how much variance is captured by leading eigenvalues \cite{marbut2023reliable};  
and finally Spectral Utilization Index (SUI), a composite metric that harmonically combines hard and soft rank to balance dominant-mode and tail usage



Through systematic analysis across the FFN width sweep $D$=$\alpha d$, where $\alpha \in \{1, 2, 2.67, 4, 5, 6, 7, 8\}$, and model sizes ranging from 70M to 250M parameters, we uncover an Asymmetric Spectral Scaling Law that fundamentally changes our understanding of capacity allocation. The power law (Log-Log) fits reveal a striking asymmetry (Figure \ref{fig:ScalingLaws_Introduction}): While soft spectral rank scales near-perfectly with FFN width ($\beta \to 1, R^2 \to 1$), hard spectral rank, measuring the dominant subspace, plateaus early with weak, noisy scaling ($\beta \approx 0.5, R^2 \approx 0.5$).

This asymmetry highlights that widening FFNs operates through {\em tail-first growth}: predominantly adding low-energy directions while the high-energy mode saturates early. In other words, capacity increases, but it is increasingly allocated to directions that carry little variance. This effect resembles the well-known spectral bias in function space, where low input frequencies are learned before high ones \cite{rahaman2019spectral}.  Both perspectives point to the same underlying principle: capacity is allocated unevenly across modes, though expressed in different bases (Fourier vs. activation eigenspectrum).

{\bf Contributions.}This work makes four main contributions:
\textbf{Conceptual.} We reframe FFN width selection, traditionally treated as an implementation detail, as a problem of spectral utilization, and introduce the first principled framework for understanding how FFN capacity is allocated with their width scaling. 
\textbf{Theoretical.} We uncover Asymmetric Spectral Scaling Laws that capture divergent growth between soft and hard spectral ranks. These laws reveal that FFN widening follows a \textit{tail-first growth} pattern, explaining why naive width scaling can yields diminishing returns.
\textbf{Methodological.} We develop a lightweight, differentiable diagnostic suite for tracking layerwise representational usage during training. This includes a closed-form estimator, $K_{\text{eff}} = 1 + (D - 1) \cdot \text{SUI}$, which links utilization to effective dimension. 
\textbf{Empirical.} Across diverse architectures and scales, we show that (i) soft/hard rank asymmetry persist across model families, (ii) optimal widths are consistently narrower than those used in practice, (iii) LayerNorm placement critically shapes utilization: Post-LN suppresses tail capacity scaling, whereas Mix-LN \cite{li2025mixln} improves dominant-mode scaling while preserving near-linear tail growth.




\section{Related Work} \label{sec:related_work}



{\bf Cost-aware neural scaling.} The foundational work \cite{kaplan2020scaling} established the power-law relations between loss and compute, later refined by the Chinchilla laws \cite{hoffmann2022an}, which showed that many models are compute-suboptimal, too wide and under-trained for their budgets. Follow-up studies \cite{sardana2023beyond} extended this perspective to deployment: under heavy traffic, the compute-optimal point shifts toward smaller models trained on more tokens, lowering inference cost. \citet{paquette2024} map the regimes where capacity, optimizer noise, or embedding quality dominate under fixed budgets. 

Other orthogonal cost factors have also been identified: vocabulary should scale with width \cite{tao2024scaling}; reduced numerical precision effectively shrinks parameter count \cite{kumar2025scaling}; and robust estimation methods enable reliable scaling-law fits from small pilot runs \cite{choshen2024hitchhiker}. These studies map efficiency trade-offs along multiple axes---compute, traffic, vocabulary, and precision. Our spectral-utilization laws introduce a {\em complementary axis:} they target latent-space usage, capturing how width is actually employed rather than measured by FLOPs alone.


\textbf{Universality and representational capacity.}
After normalizing for efficiency offsets, checkpoints spanning models from GPT-2 to PaLM have been shown to collapse onto a single sigmoidal curve, suggesting a shared scaling trajectory across architectures \cite{ruan2024observational}. The \textit{Physics of LMs} series reports a related regularity for factual knowledge: a $\leq 2$ bits/parameter ceiling that appears largely architecture-agnostic \cite{allen-zhu2025physics}. Earlier work traced such apparent universality to heavy-tailed eigenspectra and implicit self-regularization \cite{martin2021implicit}. More recent analyses refine this view: small singular values have been shown to encode critical information in pretrained Transformers \cite{staats2024locating}, while spectral collapse has been linked to over-smoothing dynamics in attention stacks \cite{dovonon2024setting}. 


{\bf Architectural and domain-specific scaling} Scaling exponents are not architecture-agnostic. \citet{tay2022scaling} show that the most effective inductive bias shifts with scale: Switch-Transformers \cite{fedus2022switch} dominate in smaller parameter regimes, Performers \cite{choromanski2020rethinking}  at mid-scale, and vanilla attention at large scale. \citet{cabannes2024scaling} derive exact scaling laws for associative-memory matrices, while \citet{shi2024scaling} explain why larger models can underperform on time-series tasks by introducing a look-back-aware law. \citet{fort2025scaling} frames adversarial robustness as a scaling phenomenon, showing that resistance to attack remains nearly constant across two orders of magnitude in model size. Finally, \citet{lyu2025a} present an analytically solvable attention mechanism that yields closed-form power laws, providing a theoretical baseline.


These threads underscore that scaling is multifaceted, bending with inductive bias, data modality, precision, and security constraints, precisely the facets our spectral scaling laws aim to highlight across GPT-2, LLaMA, and nGPT.


\section{Method}

In this section, we explain our methodology for extracting layer-wise covariance spectra from FFN internal representation, and describe the four spectral  metrics that quantify spectral utilization, and capture various aspect of spectrum (e.g., uniformity vs spikes). We finish with the end-to-end algorithm and a short complexity analysis.

\subsection{Preliminaries and Eigendecomposition}
{\bf Notation} 
Let an $L$-layer transformer be given. Each transformer consist of an FFN layer whose hidden width is $D$; the width multiplier (relative to the model’s embedding size $d$) is denoted $\alpha = D/d$. Formally, FFN with gating activation (e.g., SwiGLU in LLaMA \cite{touvron2023llama}) represented as $\text{FFN}(x) = W_{\text{down}}(\sigma(W_{\text{gate}}x) \odot (W_{\text{up}}x))$, where $\odot$ represents element-wise multiplication and $\sigma$ is activation function such as SiLU \cite{elfwing2018sigmoid}. The pre-activation (output of the first linear projection) and pos-activation (before the down-projection) is represented as $\text{PreAct}(X) = W_{\text{gate}}x$ and $\text{PostAct}(X) = \sigma((W_{\text{gate}}x) \odot (W_{\text{up}}x))$.


\begin{table*}[htbp]
\centering
\caption{Spectral utilization metrics for characterizing the FFN latent space utilization. Hard and Soft Rank capture absolute participation and entropy-based ranks in the native $[1,D]$ scale, while their normalized forms yield bounded $[0,1]$ utilization scores. Spectral concentration measures front-loading of variance, SUI balances hard and soft ranks, and eDim translates spectral patterns into an interpretable effective dimension.}
\label{tab:spectral_metrics}
\resizebox{0.99\textwidth}{!}{
\begin{tabular}{@{}lccccc@{}}
\toprule
Metric & Definition & Range & Qualitative signal & Interpretation & Cost \\
\midrule
Hard Spectral Rank & $\displaystyle \text{PR} = \frac{(\sum_i \lambda_i)^2}{\sum_i \lambda_i^2}$ , $\tilde{\text{PR}} = \frac{\text{PR} - 1}{D - 1}$ & $[0, 1]$ & Spikes $\rightarrow$ collapse & Dominant spikes & $\mathcal{O}(D)^*$ \\
\addlinespace
Soft Spectral Rank & $\displaystyle \text{eR} = \exp\left(-\sum_i p_i \log p_i\right)$ , $\displaystyle \tilde{\text{eR}} = \frac{\text{eR} - 1}{D - 1}$ & $[0, 1]$ & Long tails $\rightarrow$ dilution & Uniformity of spread & $\mathcal{O}(D)$ \\
\addlinespace
Spectral Concentration & $\displaystyle \text{SC} = \frac{2}{D} \times \sum_{k=1}^{D} \left( \frac{\sum_{i=1}^k \lambda_i}{\sum_{i=1}^D \lambda_i} - \frac{k}{D} \right)$ & $[0, 1]$ & Strength of spikes & Front-loadedness & $\mathcal{O}(D)$ \\
\addlinespace
Spectral Utilization Index & $\displaystyle \text{SUI} = \frac{2\tilde{\text{PR}}\cdot\tilde{\text{eR}}}{\tilde{\text{PR}} + \tilde{\text{eR}}}$ & $[0, 1]$ & Penalizes both extremes & Balanced utilization & $\mathcal{O}(1)^\dagger$ \\
\addlinespace
Effective dimension & $\displaystyle \text{eDim} = 1 + (D - 1)\text{SUI}$ & $[1, D]$ & \# active PCs & \# active dimensions & $\mathcal{O}(1)$ \\
\bottomrule
\multicolumn{6}{l}{\footnotesize $^*$Once eigenvalues are sorted; $^\dagger$Once ranks known} \\ 
\end{tabular}}
\end{table*}


{\bf Activation sampling and co-variance matrix formation}
During training step $t$ we sample a mini-batch of $N$ tokens from each FFN layer's ($\ell$) post-activation $X^{(\ell,t)}_{\mathrm{post}} \in \mathbb{R}^{N \times D}$. We compute the covariance using all $N$ tokens without any sub-sampling or statistical approximations to capture the true behavior of the model. Further, we compute an unbiased covariance matrix for all tokens in the batch as follows:

\vspace{-1.5em}
\begin{equation}
\Sigma = \frac{(X - \mu)^T (X - \mu)}{N - 1} \;\;\in \;\mathbb{R}^{\,D\times D}. 
\end{equation}

For each covariance matrix, we perform eigendecomposition to obtain the eigenvalues $\Sigma v = \lambda v$.  The eigenvalues are sorted in descending order: $\lambda_1 \geq \lambda_2 \geq \ldots \geq \lambda_D \geq 0$. All subsequent metrics depend only on this spectrum.

\subsection{Spectral Rank Metrics} \label{subsec:spectral_utilization_metrics}

When a feed-forward block is widened, the key question shifts from how many parameters did we add? to how many of those additional directions does the model actually use? To quantify this notion of {\em use}, we analyze the eigenspectrum of the post-activation covariance matrix and distill it into four  metrics, each lies in the range $[0,1]$ and can be computed in $\mathcal{O}(D)$ time (Table \ref{tab:spectral_metrics}).

{\bf Hard spectral rank.}
Participation Ratio (PR) acts as a hard counter of dominant directions. Since PR squares the first spectral moment and divides by the second, it is particularly sensitive to prominent eigenvalues: even a single large spike can significantly cap its value, whereas numerous smaller eigenvalues have minimal impact \cite{gao2017theory,hu2022spectrum}. Hence, PR effectively rounds off all but the strongest axes, a {\em hard} spike-sensitive estimate.

{\bf Soft Spectral Rank.} It complements PR  by measuring the Shannon entropy of the full eigenvalue distribution \cite{skean2025layer,wei2024diff,garrido2023rankme,anand2011shannon,passerini2008neumann}, by converting eigenspectrum into a  probability distributions as $p_i = \lambda_i / \sum_j \lambda_j$. Normalizing to $[0,1]$ yields a smooth measure of dimensionality that captures long-tail variance patterns. Thus, while hard rank is sensitive to dominant peaks, soft rank responds to tail behavior. Describing the pair as hard and soft therefore captures their complementary sensitivities: former reacts sharply to collapse (variance concentrated in a few axes), whereas the latter flags spectral dilution, variance diffused so widely that no direction carries significant weight.

{\bf Spectral Utilization Index} SUI combines hard and soft spectral ranks into a unified measure of spectral utilization. Hard and soft ranks independently capture opposing failure modes--spectral collapse versus dilution. To effectively combine these metrics, we adopt their harmonic mean, as it strongly penalizes imbalance: the harmonic mean sharply drops if either input is low, ensuring SUI attains high scores only when both metrics indicate balanced utilization. By rewarding spectra that avoid extremes and peak when a moderate number of principal directions carry most variance, SUI thus provides a robust, intuitive, and parameter-free indicator of overall spectral behavior. 


{\bf Spectral concentration.}
Practitioners not just about how many directions are active, but also about where the variance is concentrated. Spectral concentration measures the area between the cumulative eigen-spectrum and a uniform baseline \cite{marbut2023reliable}, where a higher value indicates that variance predominantly concentrates within the leading principal components, whereas lower value implies a more uniform distribution of variance across the spectrum. Thus, unlike previous metrics, it distinguishes spectra that utilize different fractions of the available latent space.

Finally, we convert SUI into an integer-valued measure called Effective Dimension (eDim), which directly represents the approximate number of active principal components. This makes interpretation more intuitive, particularly it simplifies abstract ratio into an absolute counts over abstract ratios and simplifies comparisons across layers of varying widths.

{\bf Why these specific metrics?}
The hard and soft ranks offer complementary perspectives on spectral utilization: one highlights spectra dominated by a few large eigenvalues, while the other captures cases with many small eigenvalues spread over a long tail. Spectral concentration metric complements these ranks by pinpointing precisely where variance accumulates. SUI unifies the two ranks into a single robust metric, penalizing both spectral extremes, and eDim further translates this into an intuitive count of active principal components. Collectively, these metrics map each layer onto an interpretable three-dimensional spectrum: collapse versus dilution, front-loaded versus dispersed variance, and overall spectral efficiency.



\begin{figure*} [t]
\centering
\subfloat[LLaMA-70M (PreLN) ]{\includegraphics[width=.33\textwidth]{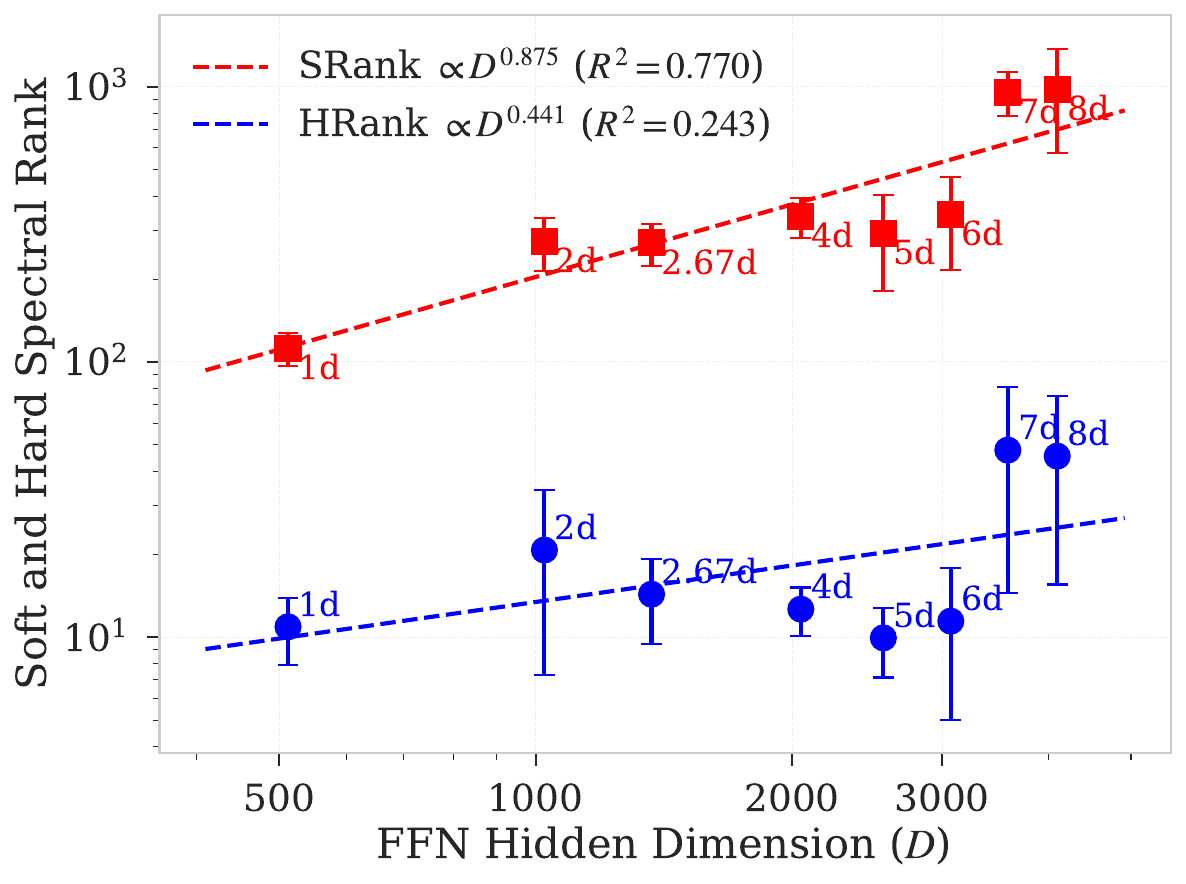}}
\subfloat[LLaMA-130M (PreLN) ]{\includegraphics[width=.33\textwidth]{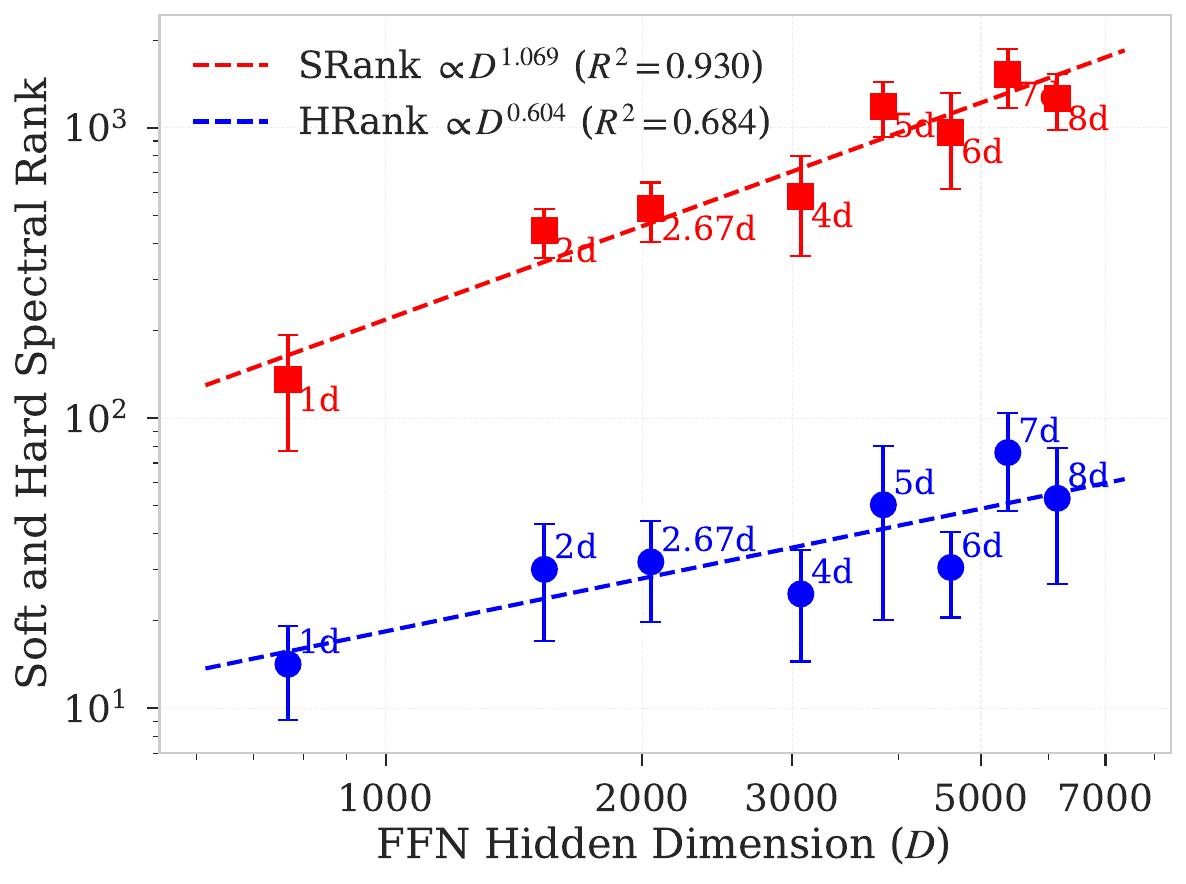}} 
\subfloat[LLaMA-250M (PreLN) ]{\includegraphics[width=.33\textwidth]{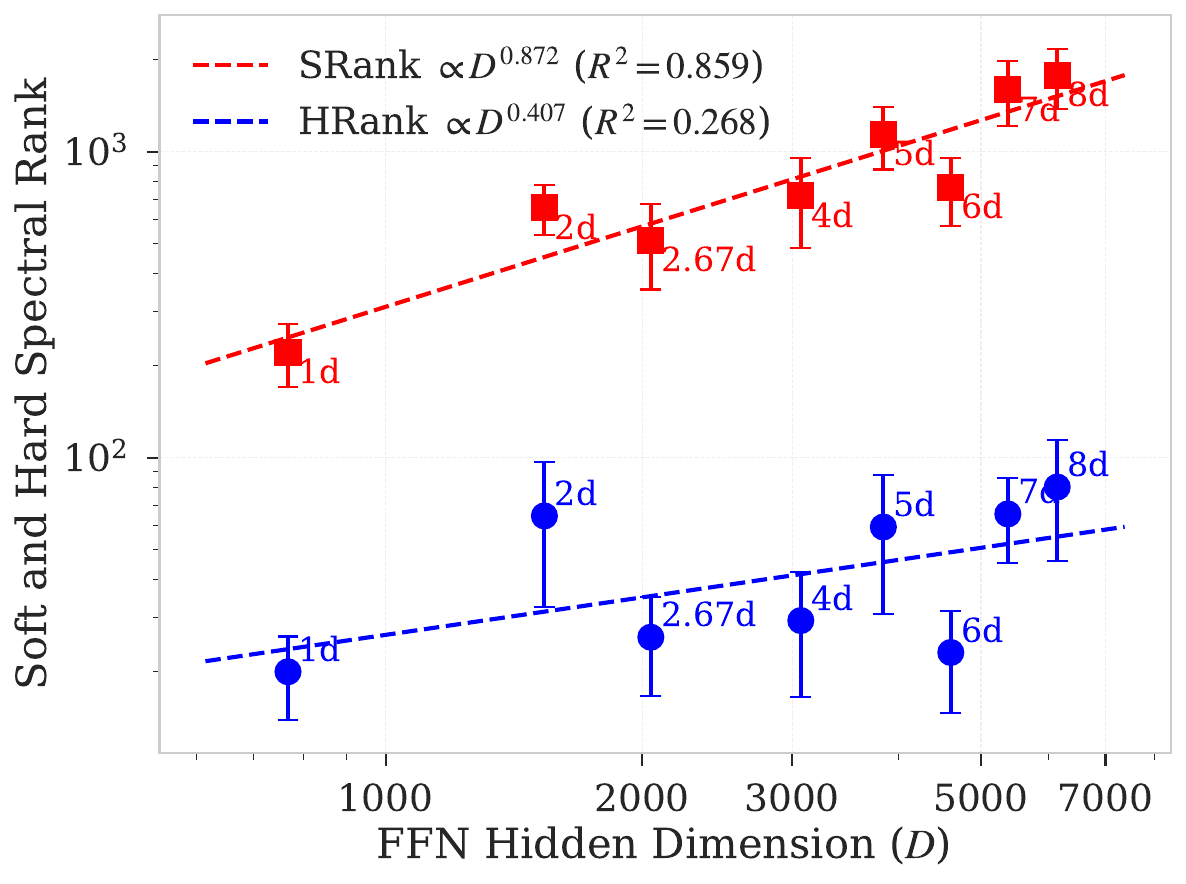}} 
\vspace{-0.5em}
\caption{ {\bf Asymmetric spectral scaling} with FFN width in LLaMA-style Pre-LN  models.
Soft rank (SRank, \textcolor{red}{\bf red}) and hard rank (HRank, \textcolor{blue}{\bf blue}) vs. FFN hidden dimension $D$ on log-log axes for (a) 70M, (b) 130M, and (c) 250M backbones (fixed $d$, width sweep $D \in \{1, 2, 2.67, 4, 5, 6, 7, 8\}$). Dashed lines are power-law fits; annotations mark $\alpha d$. Soft-rank exponents cluster near unity ($\beta = \{0.873, 1.069, 0.872\}$; $R^2 = \{0.770, 0.980, 0.850\}$), while hard-rank exponents are smaller and noisier ($\beta = \{0.441, 0.604, 0.407\}$; $R^2 = \{0.248, 0.684, 0.268\}$). All networks are trained from scratch; markers show layer median values, and error bars indicate across-layer variability.}
\label{fig:ScalingLaws_summary_all_llama_raw}
\end{figure*}

\section{Experimental Results}
In this section, we present our empirical findings on the spectral scaling laws in by varying the hidden dimension sizes of FFNs. We primarily use Hard and Soft  utilization to investigate  how each scales with the hidden dimension $D$ for three sizes of LLaMA models (70M, 130M, 250M). To study how effectively FFNs leverage increasing hidden dimensions, we trained LLaMA models from scratch on C4 datasets. For each scale, we varied the hidden dimension $D$ across 8 values, $D$=$\alpha d$, where $\alpha \in \{1, 2, 2.67, 4, 5, 6, 7, 8\}$

\subsection{Asymmetric Spectral Scaling Laws}
\textbf{Asymmetric scaling across widths.}
Across all three backbones LLaMA networks (Figure \ref{fig:ScalingLaws_summary_all_llama_raw}), the soft spectral rank follows a near-linear power law with width, whereas the hard spectral rank grows sublinearly and with greater variability. Quantitatively, SRank slopes are $\beta \approx 0.88$ (70M), $\beta \approx 1.07$ (130M), and $\beta \approx 0.87$ (250M), all with strong fits ($R^2 \approx 0.77, 0.93, 0.86$). In contrast, HRank slopes are much smaller ($\beta \approx 0.44, 0.60, 0.41$) and substantially noisier ($R^2 \approx 0.24, 0.68, 0.27$). The persistent vertical separation between SRank and HRank trends spans orders of magnitude, indicating that widening FFNs consistently inflates entropy-sensitive spectral rank  more than the core participation-ratio-defined subspace.

\textbf{Tail-first growth.}
The disparity in slopes and lower $R^2$ values for HRank point to a \textit{tail-first allocation of capacity}: as width $D$ increases, models primarily populate low-energy directions (raising SRank), while the high-energy subspace expands slowly and irregularly (limited HRank gains). The 130M case comes closest to linear SRank scaling ($\beta \approx 1.07, R^2 \approx 0.93$), yet even here the hard-rank response remains sublinear ($\beta \approx 0.60$). This asymmetry supports the interpretation that width first buys coverage of many fine-grained, low-variance modes before it substantially grows the dominant, high-variance core.

As widening predominantly enlarges the low-energy tail, returns on the dominant-mode subspace diminish with $D$. Practically, this suggests width expansion should avoid excessive tail growth, favoring tail-aware pruning (to preserve core modes and trim diffuse directions) and MoE designs that allocate experts to tail capacity rather than uniformly inflating a single dense FFN.

\subsection{Spectral Rank Utilization}

\begin{figure*} [t]
\centering
\subfloat[LLaMA-70M (PreLN) ]{\includegraphics[width=.33\textwidth]{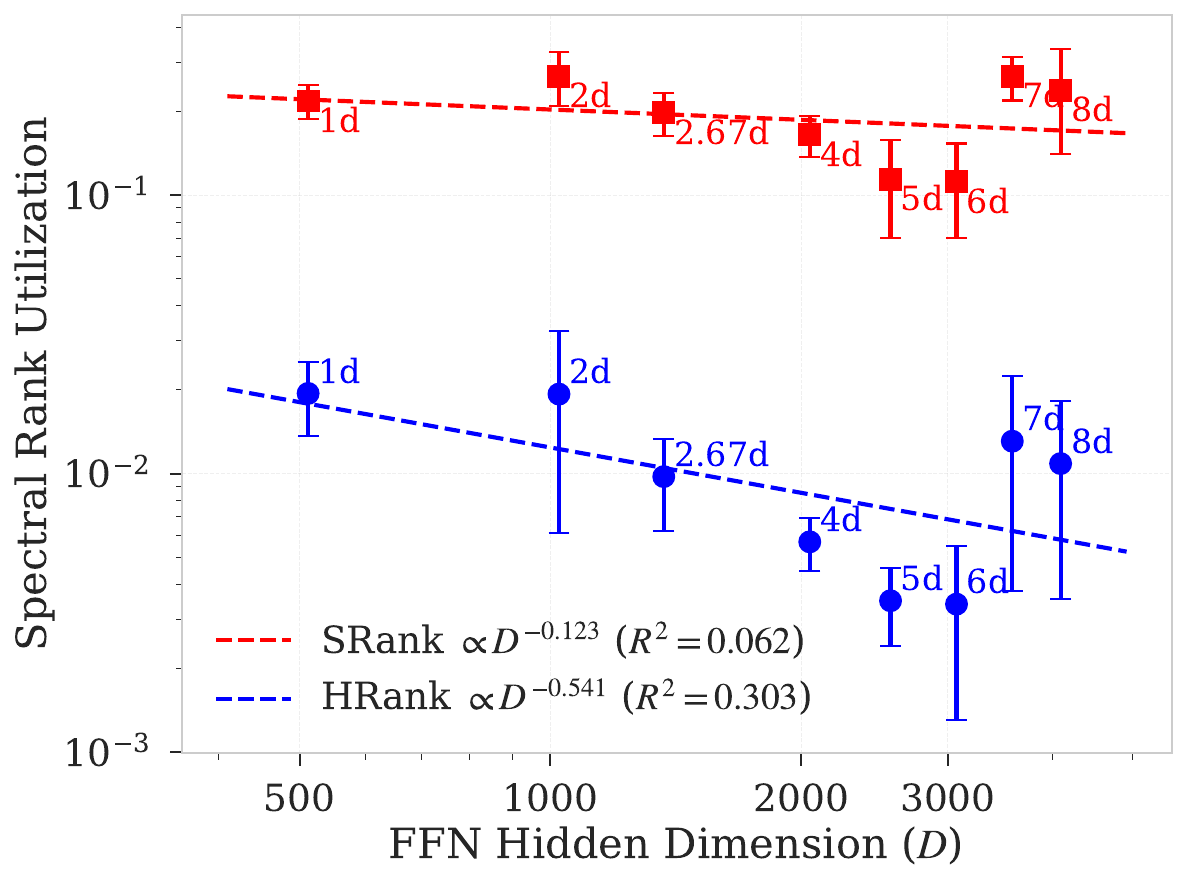}}
\subfloat[LLaMA-130M (PreLN) ]{\includegraphics[width=.33\textwidth]{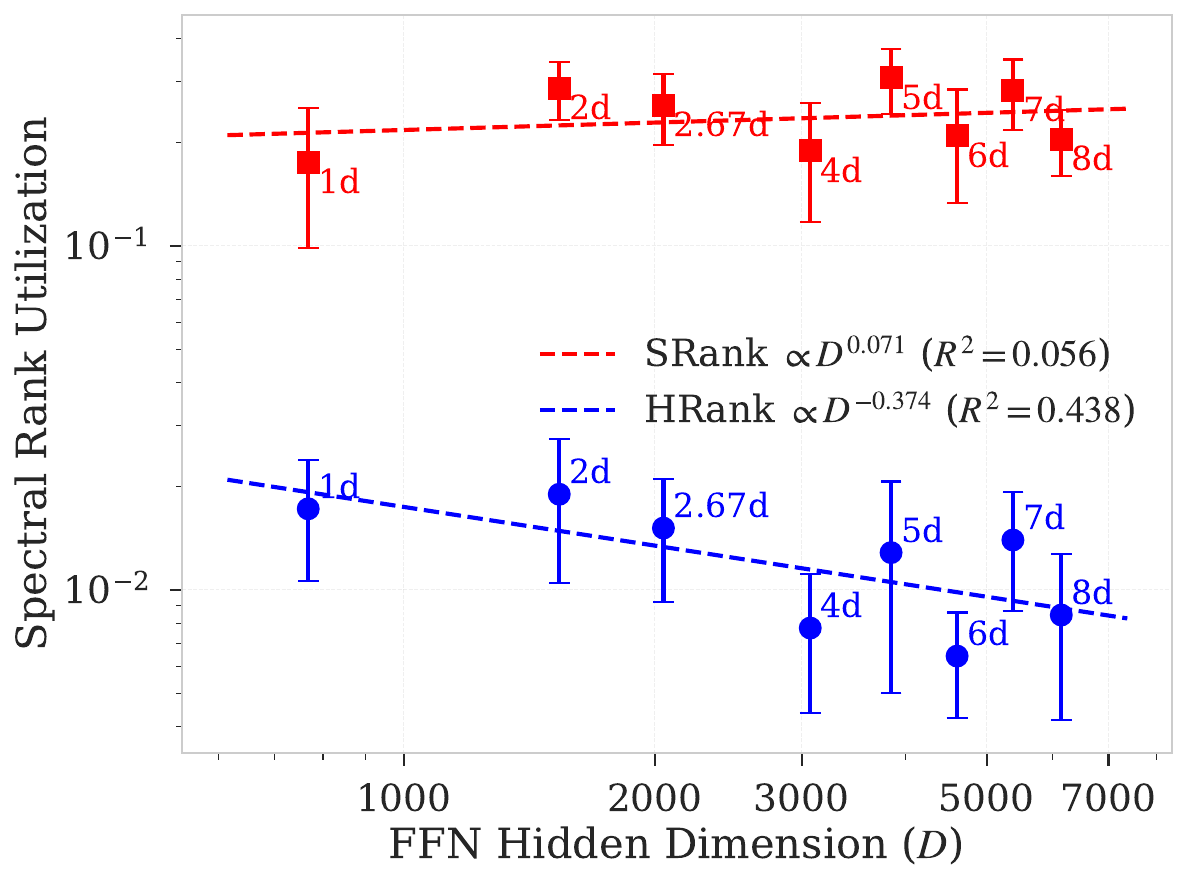}} 
\subfloat[LLaMA-250M (PreLN) ]{\includegraphics[width=.33\textwidth]{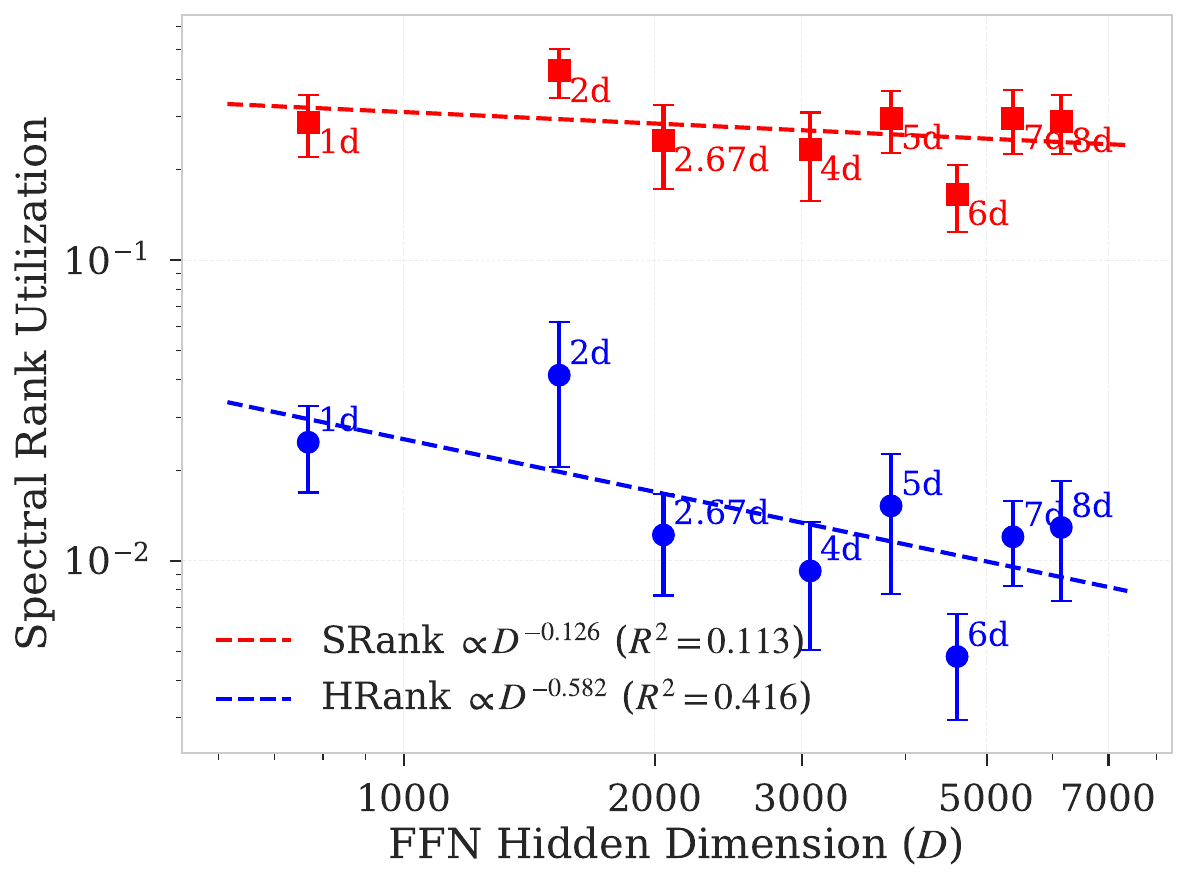}} 
\vspace{-0.5em}
\caption{{\bf Spectral-rank utilization vs. FFN width} in LLaMA-style Pre-LN models.
We plot soft-rank utilization (SRank$/(D-1)$, \textcolor{red}{\bf red}) and hard-rank utilization (HRank$/(D-1)$, \textcolor{blue}{\bf blue}) vs. FFN hidden dimension $D$ on log-log axes for 70M, 130M, and 250M backbones (fixed depth; width sweep $D = \alpha d, \alpha \in \{1, 2, 2.67, 4, 5, 6, 7, 8\}$). Dashed lines show power-law fits, highlighting that SRank scales nearly linearly with width while HRank grows more slowly and with higher variability. All networks are trained from scratch; markers indicate layer median, and error bars denote across-layer variability.}
\label{fig:ScalingLaws_summary_all_llama}
\end{figure*}

\textbf{From capacity to efficiency.}
Normalizing ranks by $D$ turns them into utilization fractions, $\tilde{HR}$ and $\tilde{SR}$. Across scales, $\tilde{HR}$ declines reliably with width, confirming that the high-energy mode occupies a shrinking share of dimensions as $D$ grows (e.g., slopes around $-0.5$ across 70M/130M/250M). By contrast, $\tilde{SR}$ is nearly \textit{scale-invariant} (slopes $\approx 0$), showing that the low-energy tail keeps pace with widening.

\textbf{Consistency with the asymmetric law.}
Algebraically, if SRank $\propto D^{\beta_{\text{soft}} \sim 1}$ and HRank $\propto D^{\beta_{\text{hard}} < 1}$, then $\frac{\text{SRank}}{D} \propto D^{\beta_{\text{soft}} - 1} \approx D^0$ and $\frac{\text{HRank}}{D} \propto D^{\beta_{\text{hard}} - 1} \downarrow$, exactly matching the observed near-flat soft utilization and negative hard utilization slopes. Put simply, widening allocates capacity \textit{tail-first}: coverage expands, but the fraction devoted to the core contracts.


\begin{table*}
\centering
\caption{Effective dimension (eDim), shown in gray-shaded columns, together with hard- and soft-rank spectral metrics for LLaMA models under width scaling ($D = \alpha d$, $\alpha \in {1, 2, 2.67, 4, 5, 6, 7, 8}$), where $d$ is the model embedding dimension (512 for 70M; 768 for 130M and 250M).} \vspace{-1em}
\label{tab:llama_summary_effective_dims}
\resizebox{0.99\textwidth}{!}{
\setlength{\tabcolsep}{3pt}
\begin{tabular}{l*{24}{c}}
\toprule
& \multicolumn{3}{c}{D=1d} & \multicolumn{3}{c}{D=2d} & \multicolumn{3}{c}{D=2.67d} & \multicolumn{3}{c}{D=4d} & \multicolumn{3}{c}{D=5d} & \multicolumn{3}{c}{D=6d} & \multicolumn{3}{c}{D=7d} & \multicolumn{3}{c}{D=8d} \\
\cmidrule(lr){2-4} \cmidrule(lr){5-7} \cmidrule(lr){8-10} \cmidrule(lr){11-13} \cmidrule(lr){14-16} \cmidrule(lr){17-19} \cmidrule(lr){20-22} \cmidrule(lr){23-25}
& HRank & SRank & \cellcolor{gray!20}{\bf eDim} & HRank & SRank & \cellcolor{gray!20}{\bf eDim} & HRank & SRank & \cellcolor{gray!20}{\bf eDim} & HRank & SRank & \cellcolor{gray!20}{\bf eDim} & HRank & SRank & \cellcolor{gray!20}{\bf eDim} & HRank & SRank & \cellcolor{gray!20}{\bf eDim} & HRank & SRank & \cellcolor{gray!20}{\bf eDim} & HRank & SRank & \cellcolor{gray!20}{\bf eDim} \\
\midrule
70M & 11 & 112 & \cellcolor{gray!20}19 & 21 & 274 & \cellcolor{gray!20}38 & 14 & 271 & \cellcolor{gray!20}26 & 13 & 338 & \cellcolor{gray!20}24 & 10 & 293 & \cellcolor{gray!20}18 & 11 & 344 & \cellcolor{gray!20}21 & 48 & 955 & \cellcolor{gray!20}90 & 46 & 975 & \cellcolor{gray!20}86 \\
130M & 14 & 135 & \cellcolor{gray!20}25 & 30 & 442 & \cellcolor{gray!20}56 & 32 & 525 & \cellcolor{gray!20}56 & 25 & 582 & \cellcolor{gray!20}47 & 50 & 1184 & \cellcolor{gray!20}96 & 31 & 964 & \cellcolor{gray!20}58 & 76 & 1521 & \cellcolor{gray!20}144 & 53 & 1257 & \cellcolor{gray!20}101 \\
250M & 20 & 221 & \cellcolor{gray!20}36 & 65 & 655 & \cellcolor{gray!20}117 & 26 & 514 & \cellcolor{gray!20}49 & 29 & 717 & \cellcolor{gray!20}56 & 59 & 1136 & \cellcolor{gray!20}112 & 23 & 764 & \cellcolor{gray!20}44 & 66 & 1593 & \cellcolor{gray!20}125 & 80 & 1777 & \cellcolor{gray!20}153 \\
\bottomrule
\end{tabular}}
\end{table*}

\textbf{Failure modes in utilization space.}
This view cleanly separates two regimes. \textit{Spectral dilution} arises when normalized soft Spectral Rank remains flat (or slightly increasing) while the normalized soft spectral rank falls, clearly noticeable in LLaMA-130M. \textit{Spectral collapse} appears when both utilizations decrease, pronounced at large $D$ for 250M. These patterns are consistent across backbones and independent of absolute width, making them a compact efficiency diagnostic.



\begin{figure*} [t]
\centering
\subfloat[$D$ = 768\label{subfig:eee_768}]{\includegraphics[width=.33\textwidth]{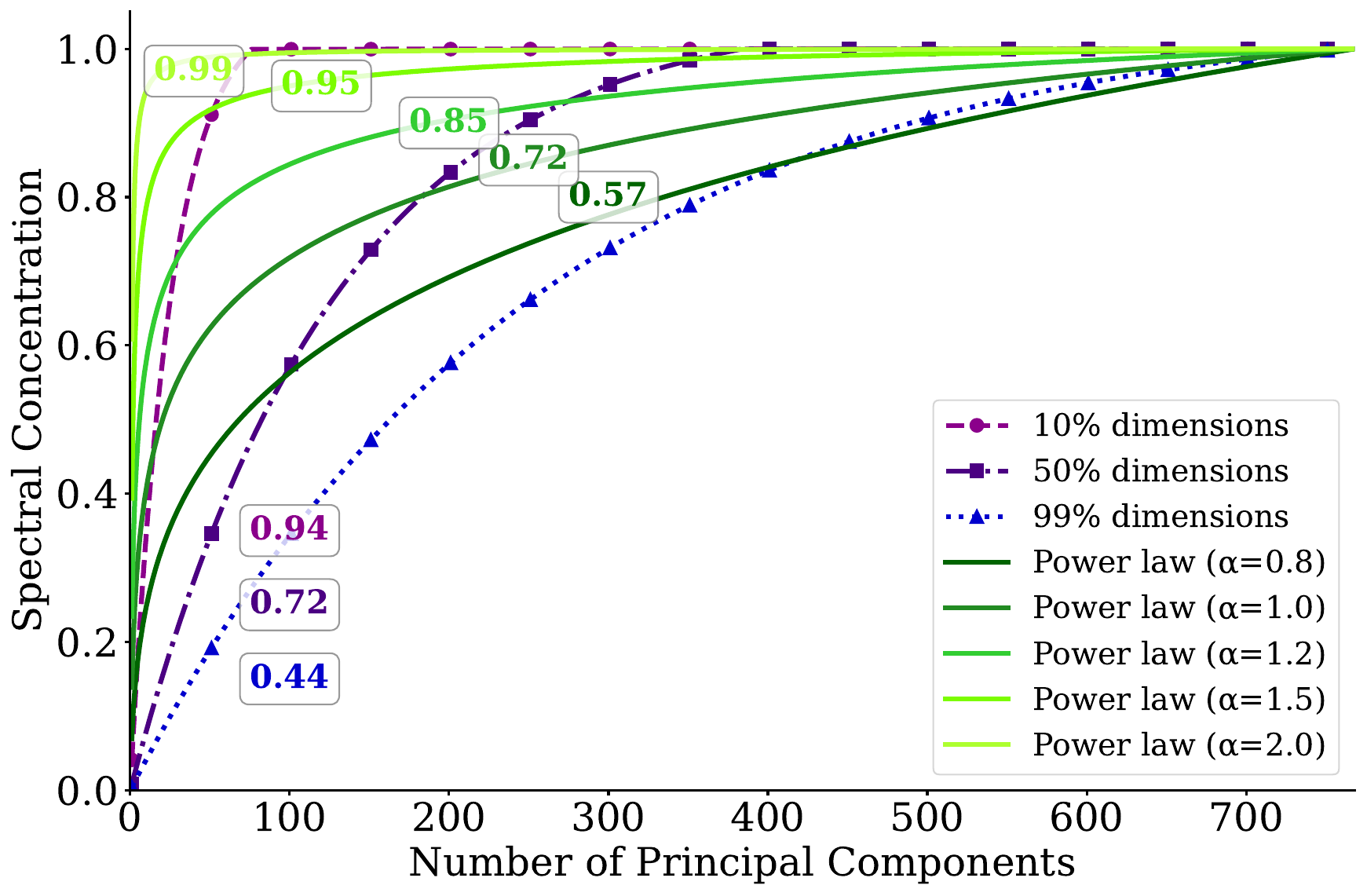}}
\subfloat[$D$ = 2048 \label{subfig:eee_2048}]{\includegraphics[width=.33\textwidth]{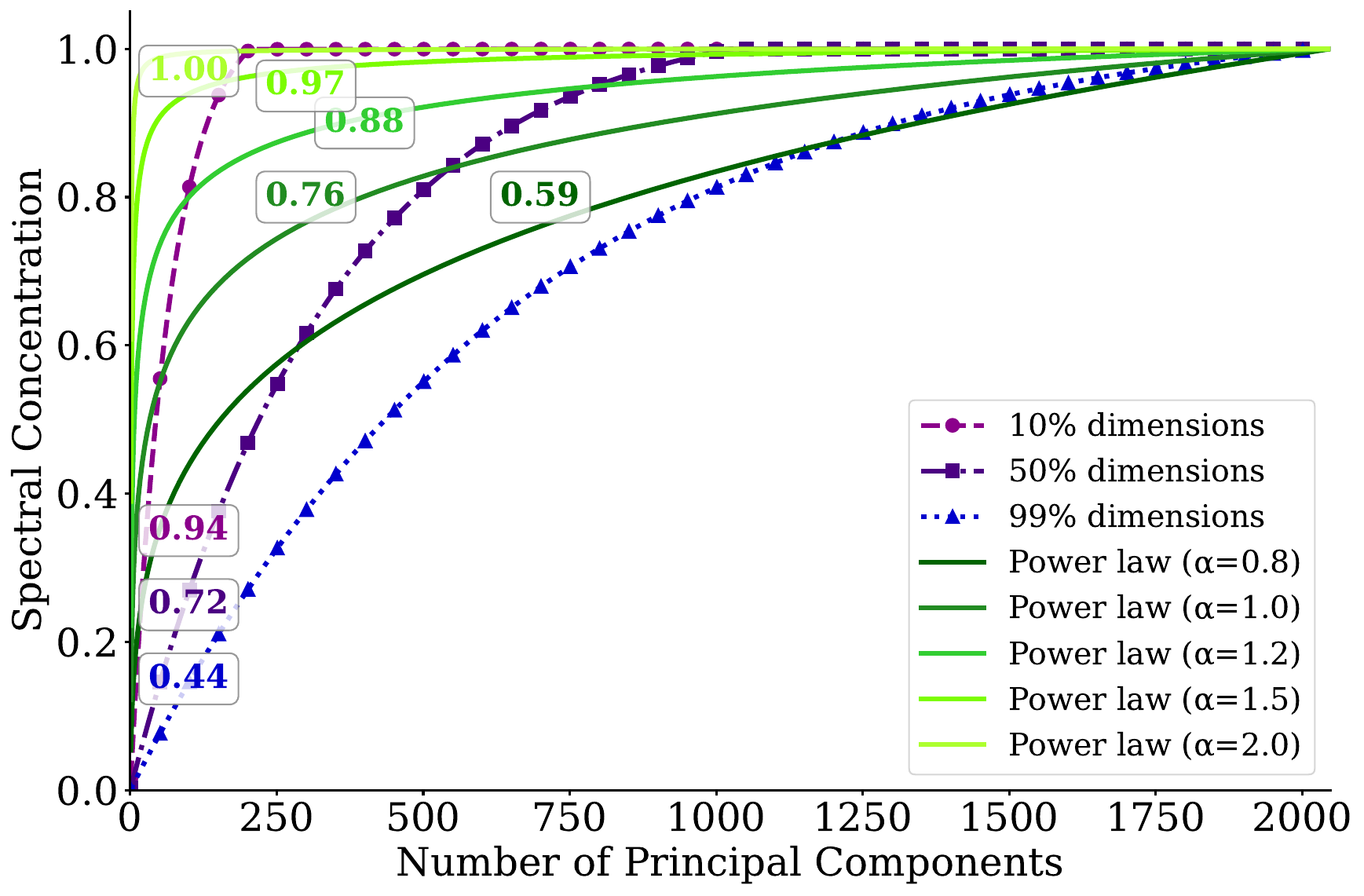}} 
\subfloat[$D$ = 3072 \label{subfig:eee_3072}]{\includegraphics[width=.33\textwidth]{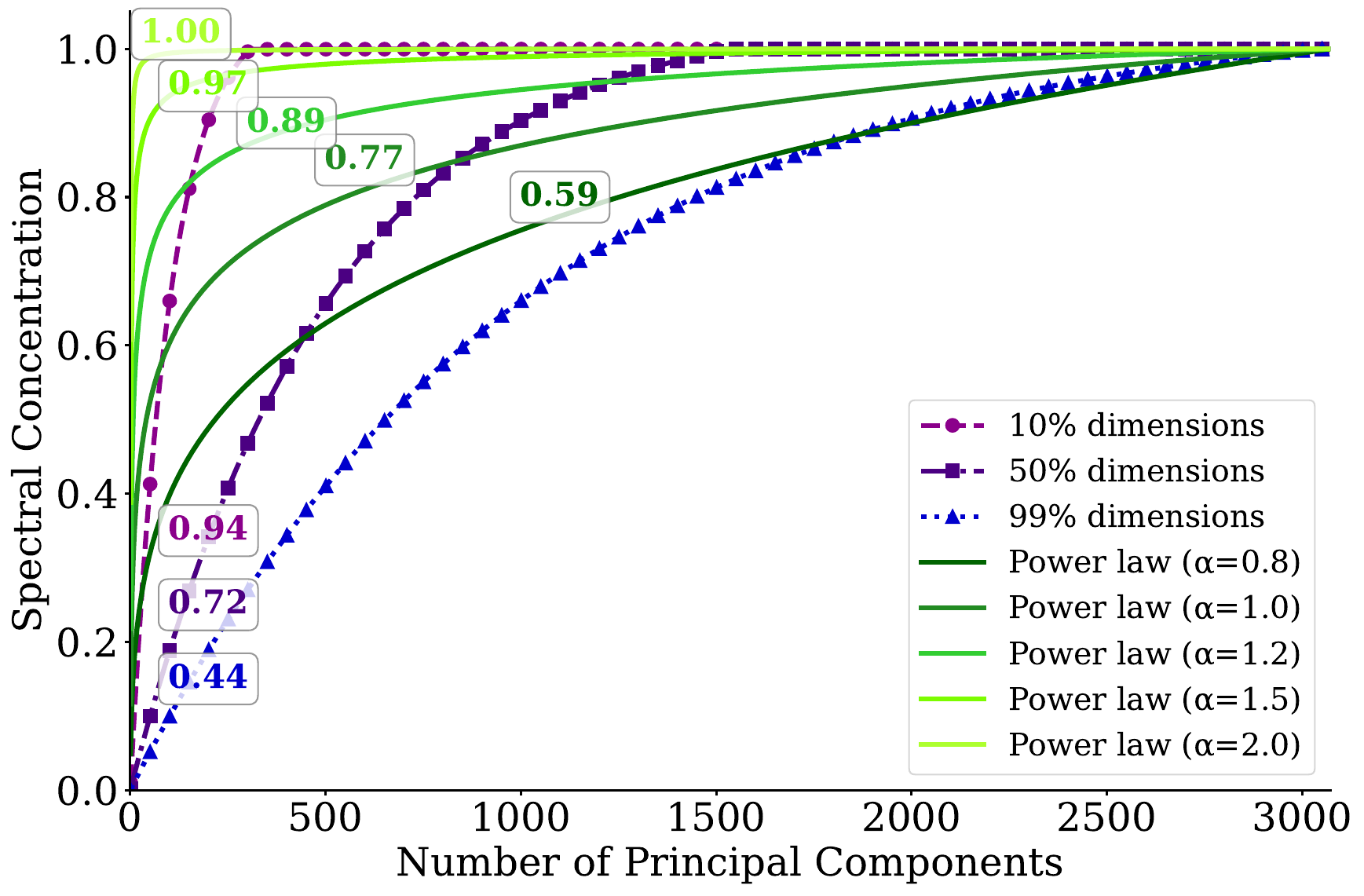}} 
\vspace{-0.5em}
\caption{Power-law templates for spectral concentration. Cumulative-variance curves generated from synthetic power-law spectra $\lambda_k \propto k^{-\alpha}$ for three latent sizes $(D=768, 2048, 3072)$. Larger exponents ($\alpha$) front-load variance and push the curve upward. Coloured call-outs report the concentration value reached by benchmark cut-offs. 
}
\label{fig:EEE_PowerLaws}
\end{figure*}

\begin{table*}
\centering
\caption{Quantitative summary of the curves in Fig \ref{fig:EEE_PowerLaws}. For each $\alpha$ and hidden size $D$ we list the variance carried by the top-1 eigenvalue, and cumulative variance captured by the first 10\%, 25\% and 50\% principal components, along with the concentration score. The results show sharp transition around $\alpha \approx 1.2$: below it at least half the spectrum is needed to explain 80\% of the variance (dilution), above it fewer than 10\% directions suffice (collapse).
} \vspace{-1em}
\renewcommand{\arraystretch}{1.2}
\small
\resizebox{0.99\textwidth}{!}{
\begin{tabular}{cccccccccccccccc}  \toprule
\multirow{2}{*}{$\alpha$} & \multicolumn{3}{c}{Top-1 eigenvalue} & \multicolumn{3}{c}{Variance @ 10\% dimensions} & \multicolumn{3}{c}{Variance @ 25\% dimensions} & \multicolumn{3}{c}{Variance @ 50\% dimensions} & \multicolumn{3}{c}{Spectral Concentration} \\
 \cmidrule(lr){2-4} \cmidrule(lr){5-7} \cmidrule(lr){8-10}  \cmidrule(lr){11-13}  \cmidrule(lr){14-16}
& 768 & 2048 & 3072 & 768 & 2048 & 3072 & 768 & 2048 & 3072 & 768 & 2048 & 3072 & 768 & 2048 & 3072 \\  \toprule
0.8 & 6.9\% & 5.4\% & 4.9\% & 51.9\% & 54.3\% & 55.2\% & 68.4\% & 70.0\% & 70.5\% & 83.1\% & 84.0\% & 84.3\% & 0.57 & 0.59 & 0.59 \\
1.0 & 13.8\% & 12.2\% & 11.6\% & 68.2\% & 72.0\% & 73.3\% & 80.8\% & 83.1\% & 83.9\% & 90.4\% & 91.6\% & 91.9\% & 0.72 & 0.76 & 0.77 \\
1.2 & 23.4\% & 22.2\% & 21.8\% & 81.9\% & 85.9\% & 87.2\% & 90.1\% & 92.3\% & 93.0\% & 95.4\% & 96.4\% & 96.7\% & 0.85 & 0.88 & 0.89 \\
1.5 & 39.4\% & 38.9\% & 38.8\% & 93.9\% & 96.3\% & 97.0\% & 97.2\% & 98.3\% & 98.6\% & 98.8\% & 99.3\% & 99.4\% & 0.95 & 0.97 & 0.97 \\
2.0 & 60.8\% & 60.8\% & 60.8\% & 99.3\% & 99.7\% & 99.8\% & 99.8\% & 99.9\% & 99.9\% & 99.9\% & 100.0\% & 100.0\% & 0.99 & 1.00 & 1.00 \\
\bottomrule
\end{tabular}}
\label{tab:eigenvalue_metrics}
\end{table*}

{\bf Composite diagnostics. }
Hard rank reflects the dominant modes and stays relatively flat (lower) across width while soft rank tracks the tail and grows steadily. Each metric alone can be {\em misleading:} soft rank continues to grow even when dominant modes are saturated, while hard rank ignores meaningful tail growth. Our notion of 
effective dimension (eDim), a harmonic-mean fusion, penalizes this imbalance and increases only when both dominant and tail capacities improve. 
As shown in Table \ref{tab:llama_summary_effective_dims}, eDim grows sub-linearly with width and remains a small fraction of $D$S for all models, underscoring the asymmetry: widening mainly expands the tail while dominant modes saturate early. Larger models achieve slightly higher eDim at the same width multiplier, suggesting better tail utilization, however,  still far from proportional scaling.

{\bf Implications for model design.}
Our results suggest a simple, spectral rationale for common FFN width choices. Since hard rank saturates early and soft rank keeps growing, the marginal eDim gain per unit width drops beyond $\sim 2.67\times$-$4\times$. LLM families that target stronger tail expressivity (e.g., GPT-2) may push to $4\times$, while those prioritizing parameter efficiency (e.g., LLaMA) can stop nearer $2.67\times$ without losing dominant-mode capacity. This is one plausible factor (among data, depth, training recipe) behind the observed widths.

From an FFN design perspective, this spectral view also yields a practical rule of thumb and a general diagnostic. By monitoring eDim during training, one can detect when widening ceases to provide meaningful returns. If eDim plateaus while hard rank remains flat, so that eDim/$D$ stagnates, dominant modes are saturated and further width only inflates tail capacity. At that point, it is more effective to freeze width and reallocate budget (e.g., to depth) or pursue layer-wise adjustments, rather than continue uniform widening.

Finally, our spectral analysis also informs pruning and layer-wise adaptation. Layers with persistently low eDim at large $D$ are natural candidates for FFN pruning or width reduction, whereas layers where eDim continues to rise with $D$ can absorb additional width more effectively. This motivates non-uniform width allocation across depth, pruning or narrowing saturated layers while widening those that remain expressive, rather than blanket scaling.

\subsection{Scaling Laws for Spectral Concentration}

We investigate the spectral concentration of FFNs activation covariance matrices by modeling their eigenvalue distribution via a truncated power-law:
$\lambda_k \propto k^{-\alpha}, \quad k = 1,\dots,D,$ where the exponent $\alpha$ controls how variance is distributed across eigen-directions. While traditional rank-based metrics (e.g., Hard and Soft Spectral Ranks) integrate information from \emph{all} eigenvalues, they often overlook crucial details in the distribution's shape, such as distinguishing between sharply peaked spectra with extensive flat tails and those smoothly decaying. The proposed power-law scaling framework directly addresses this limitation, isolating the shape characteristics of spectral distributions. Higher values of $\alpha$ yield spectra sharply concentrated (front-loaded) among leading directions, indicating incipient collapse, whereas lower values produce more uniform (diluted) distributions, indicative of suboptimal variance allocation (Fig. \ref{fig:EEE_PowerLaws}).

Empirically, several robust trends emerge from our analysis. Spectral concentration, monotonically increases with $\alpha$: as $\alpha$ rises from 0.8 to 2.0, it grows consistently from around $0.57$ to nearly $0.99$ (Table \ref{tab:eigenvalue_metrics}). Once eigenvalues decay faster than $k^{-2}$, variance is predominantly concentrated in the initial directions, becoming effectively dimension-invariant and independent of model width. This invariance enables meaningful comparisons of FFN efficiency across models of different sizes by aligning them on a common spectral utilization axis.


\begin{figure*} [t]
\centering
\subfloat[{\scriptsize Hard Rank $\beta$ evolution}]{\includegraphics[width=.25\textwidth]{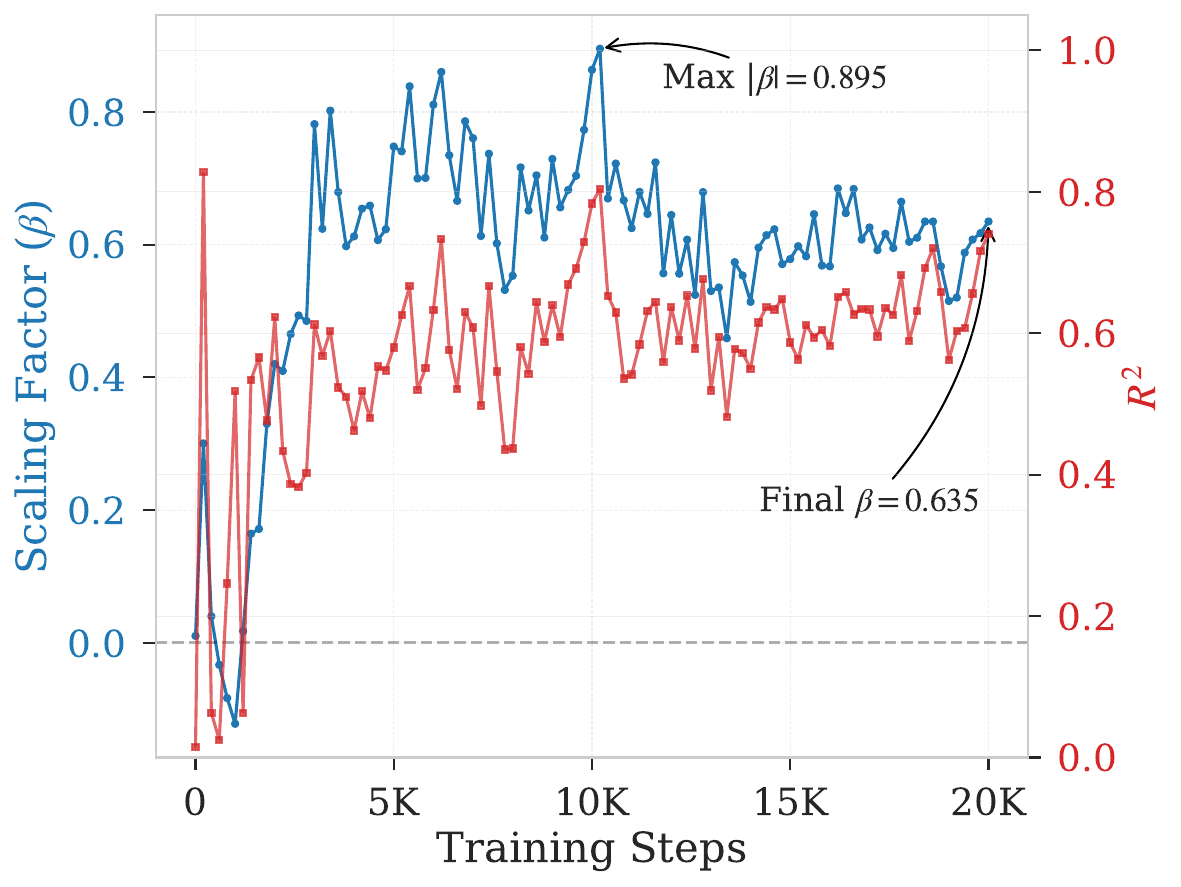}}
\subfloat[{\scriptsize Soft Rank $\beta$ evolution}]{\includegraphics[width=.25\textwidth]{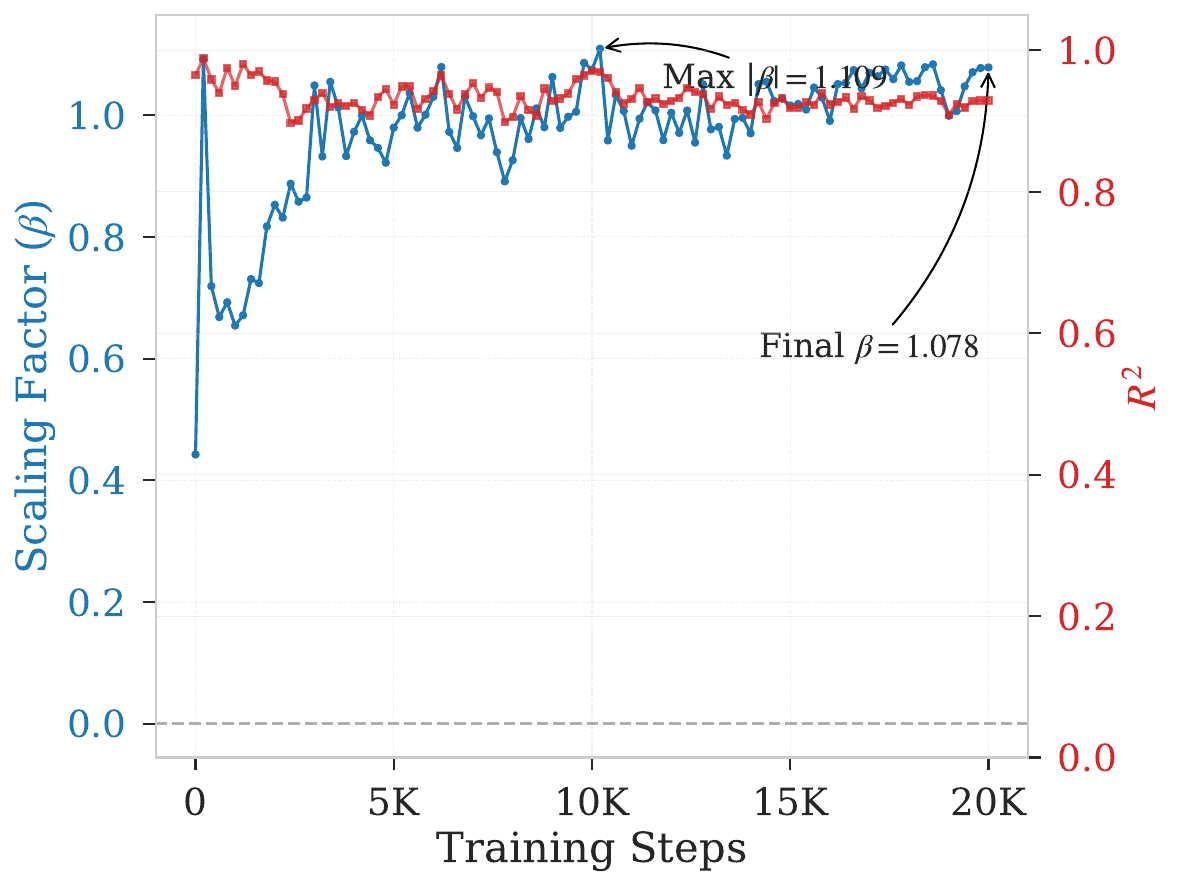}} 
\subfloat[{\scriptsize Hard Rank  Dynamics}]{\includegraphics[width=.25\textwidth]{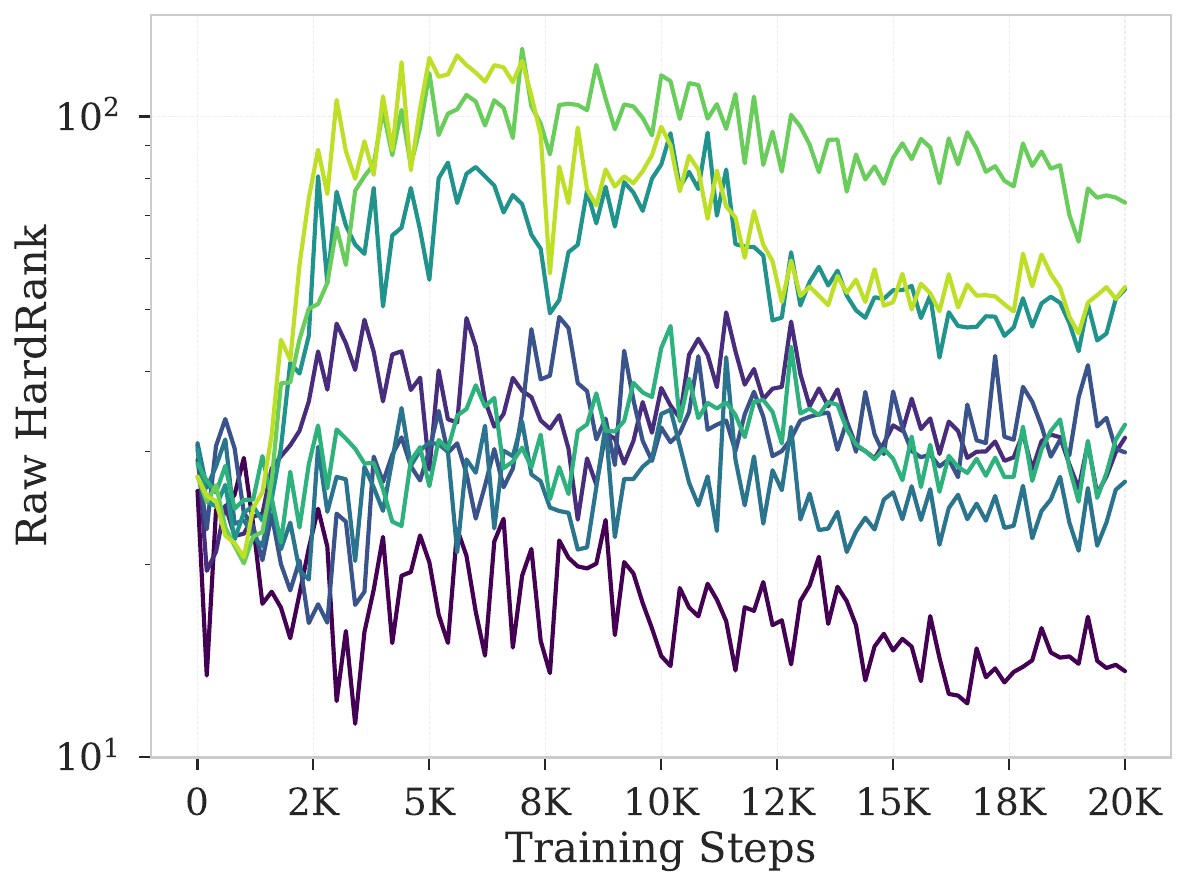}} 
\subfloat[{\scriptsize Soft Rank Dynamics}]{\includegraphics[width=.25\textwidth]{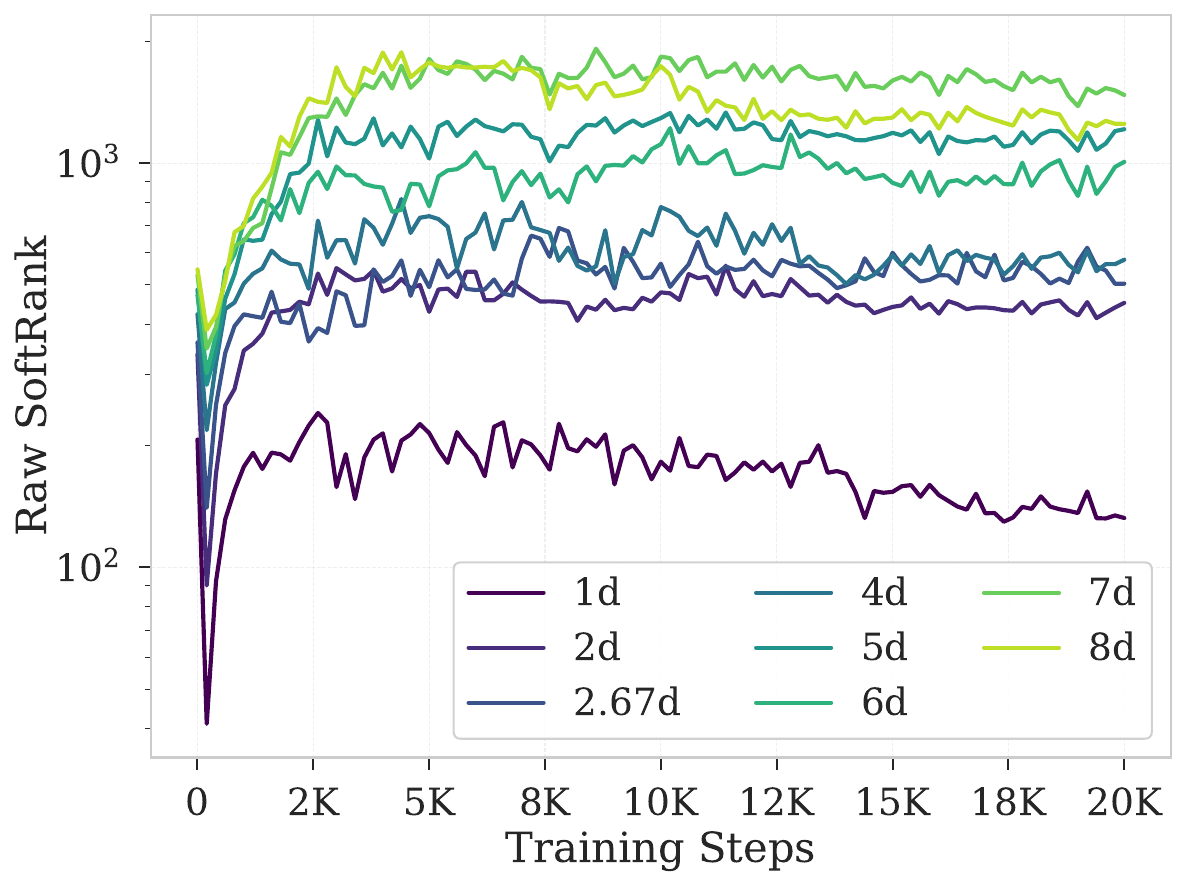}} \\ \vspace{-0.6em}
\subfloat[{\scriptsize Hard Rank Utilization ($\beta$)}]{\includegraphics[width=.25\textwidth]{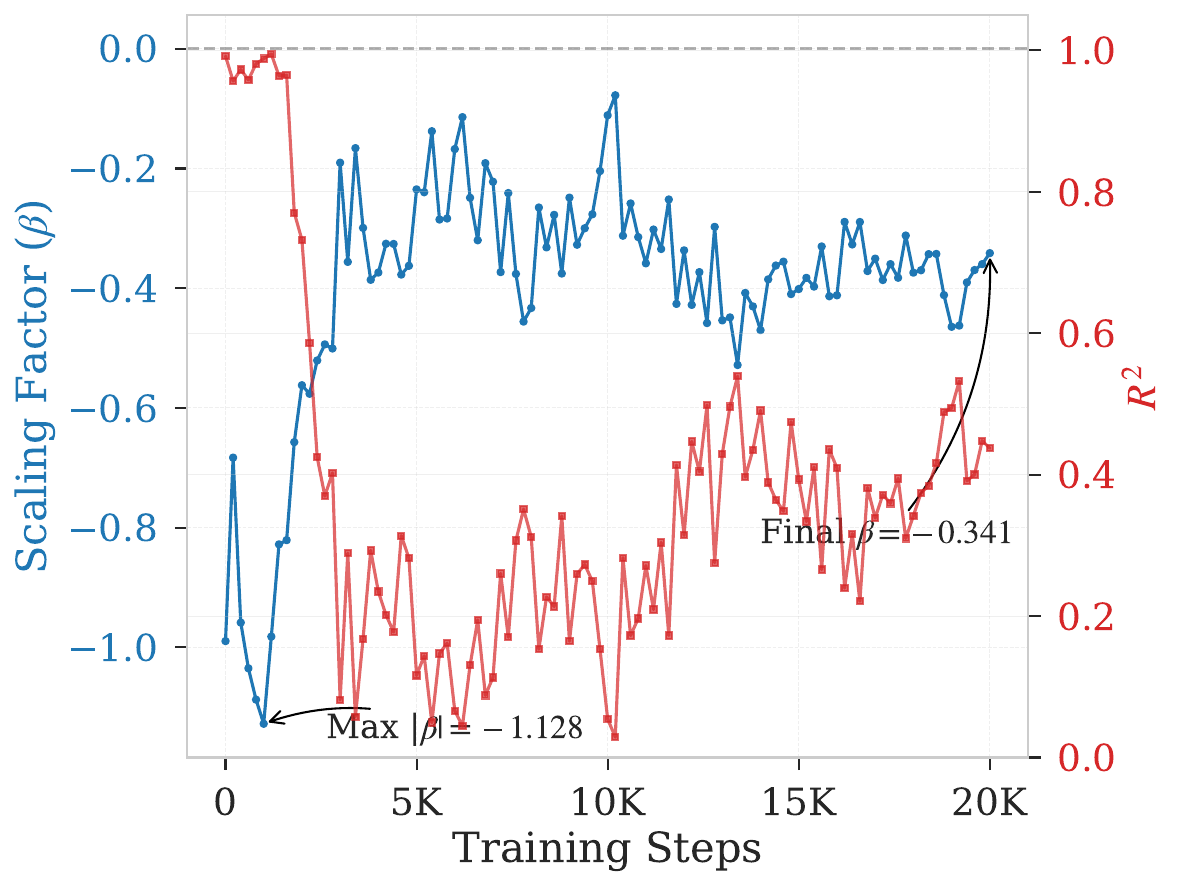}}
\subfloat[{\scriptsize Soft Rank Utilization ($\beta$)}]{\includegraphics[width=.25\textwidth]{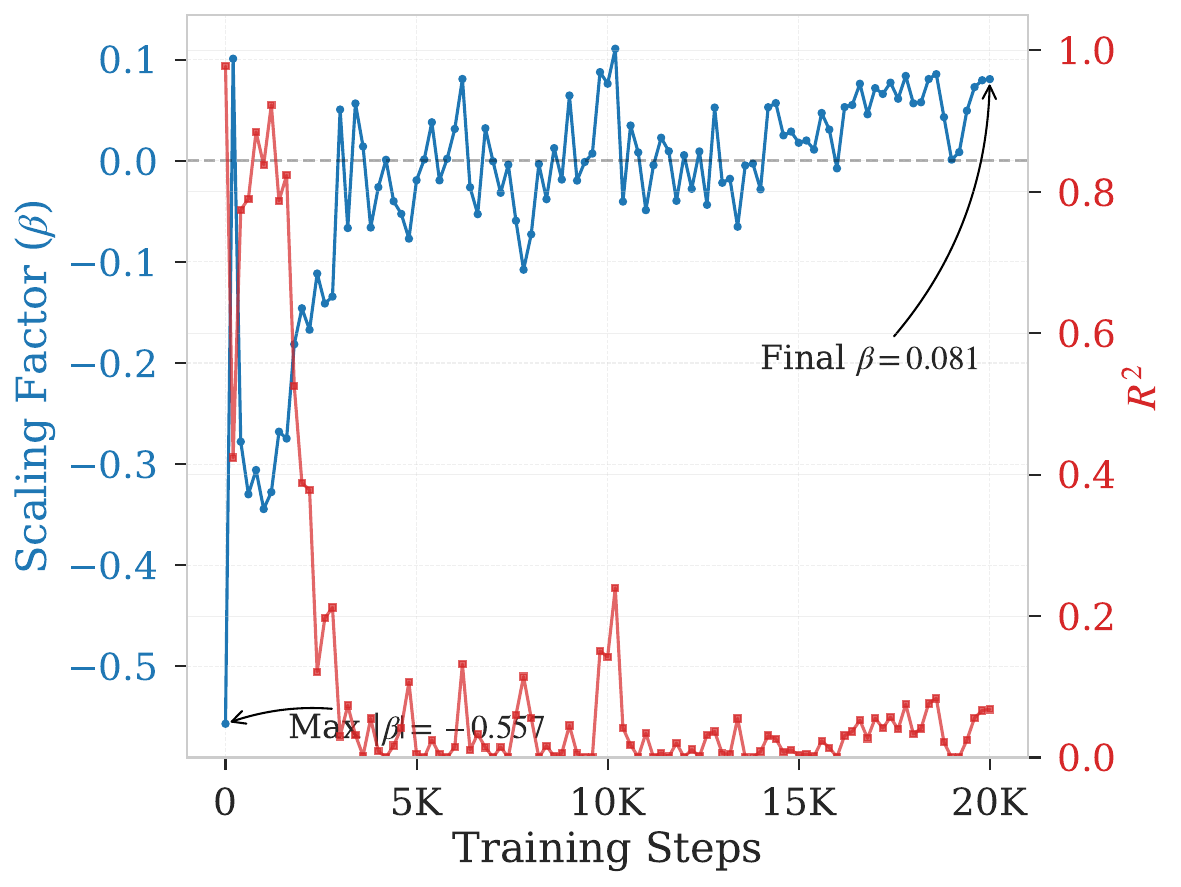}} 
\subfloat[{\scriptsize  Hard Rank Utilization Dynamics}]{\includegraphics[width=.25\textwidth]{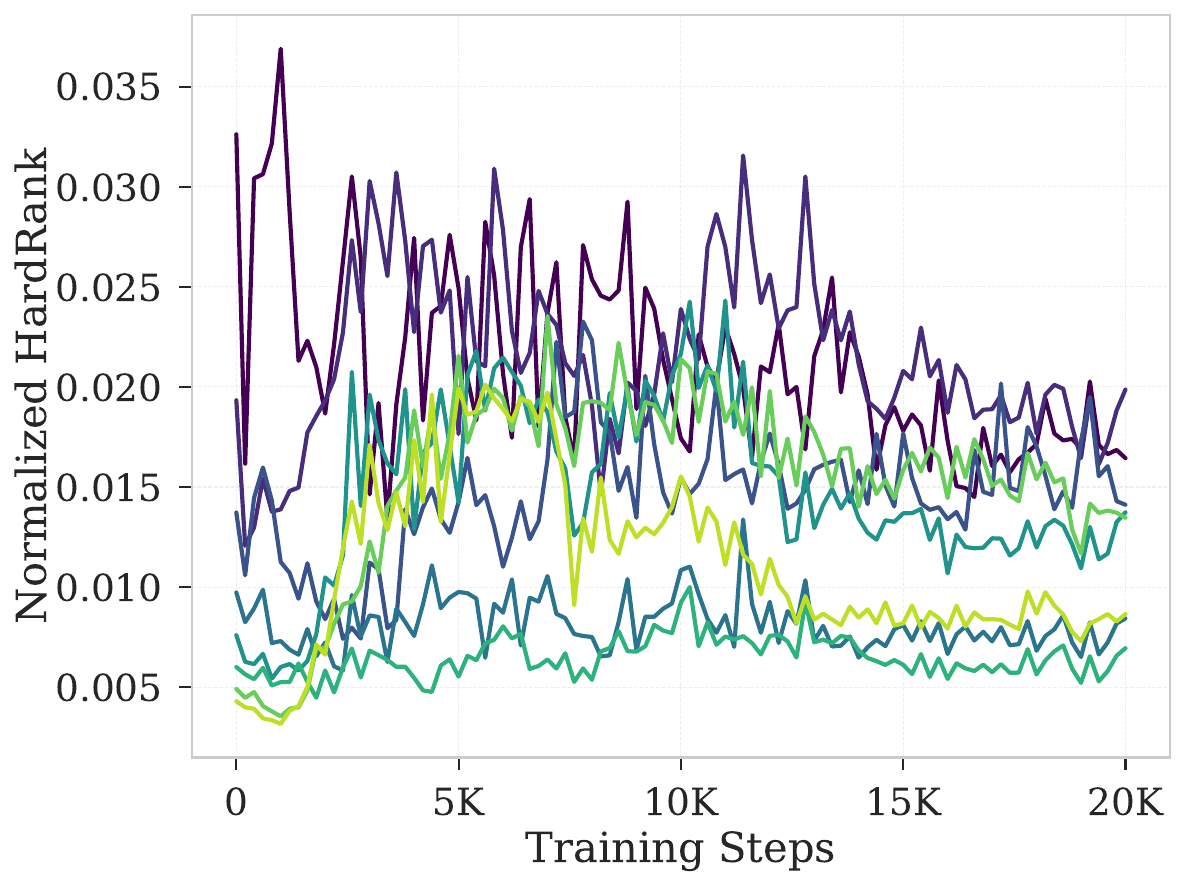}} 
\subfloat[{\scriptsize Soft Rank Utilization Dynamics}]{\includegraphics[width=.25\textwidth]{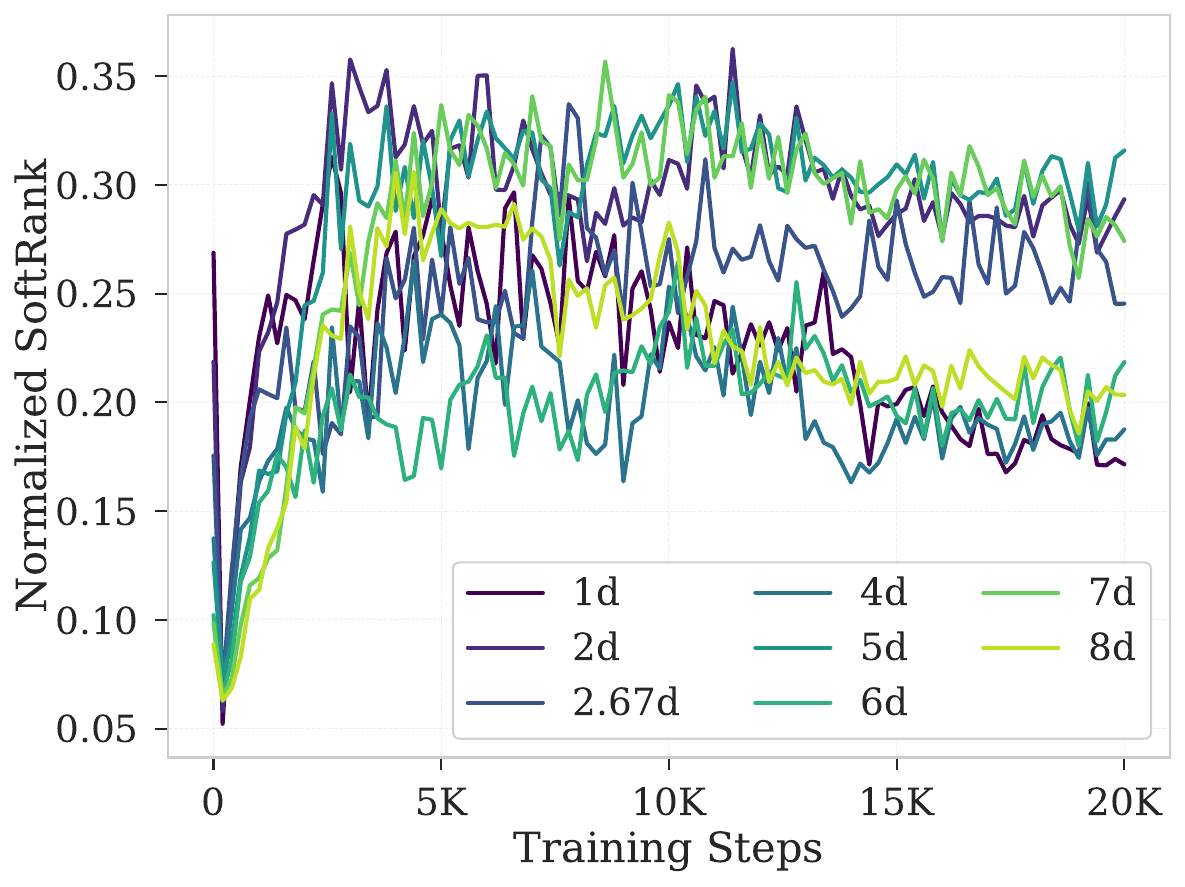}} \\ \vspace{-0.5em}
\caption{Training-time evolution of spectral scaling laws for LLaMA-130M (PreLN). Upper panels (a-d) show raw Hard- and Soft-Rank, while lower panels (e-h) illustrate normalized ranks (Rank utilization). (a,b) and (e,f) track the scaling exponent $\beta$ (\textcolor{blue}{blue}, left axis) and fit quality $R^2$ (\textcolor{red}{red}, right axis), while (c,d) and (g,h) show the corresponding layer-averaged rank dynamics fo each FFN widths (D=1$d$ to 8$d$). }
\label{fig:ScalingLaws_dynamics_beta_llama130m}
\end{figure*}

For larger $\alpha \geq 1.5$, over 90\% of variance resides within merely the top 10\% of principal components (Table \ref{tab:eigenvalue_metrics}). Conversely, at smaller values ($\alpha \approx 0.8$), capturing the same variance requires more than 50\% of components, leading to a state we term {\em spectral dilution}. Notably, activations in prevalent models such as LLaMA typically exhibit intermediate spectral concentration ($\alpha \approx 1.1$--$1.3$), thereby balancing effective dimensionality and representational compactness, avoiding the extremes of either spectral dilution or collapse.

\subsection{Spectral Scaling Dynamics}

We track rank-width behavior throughout training to ensure whether scaling relations hold reliably and to disentangle transient artifacts from persistent effects. Our aim is to pinpoint when a stable power-law regime emerges and to distinguish between FFN capacity growth (unnormalized ranks) and width efficiency(normalized ranks).

In the early phase (${\approx}2$-$3$K steps), both Hard- and Soft-Rank increases with width, but their trajectories diverge (see Figure \ref{fig:ScalingLaws_dynamics_beta_llama130m}). Hard-Rank is noisy, reflecting sensitivity to the top singular directions, whereas Soft-Rank increases smoothly and stabilizes earlier as it aggregates contributions across the spectrum. By ${\approx}5$K steps, the scaling curves flatten, $R^2$ exceeds $0.6$, and a consistent power-law regime emerges. Notably, width ordering is not {\em strictly} preserved in raw ranks: occasional crossovers occur at higher widths (more visibly in Hard-Rank than Soft-Rank), indicating transient re-allocation of capacity at higher across widths.

Normalized spectral ranks (utilization) stabilize in terms of exponents with $\beta_{\text{hard}} \approx -0.34$ and $\beta_{\text{soft}} \approx +0.08$ (final 1K steps). This implies that increasing width reduces dominant-mode concentration (lower Hard utilization) and spreads mass across more directions (higher Soft utilization). However, width ordering in the normalized curves is also not reliably preserved after $\approx$5K steps: Hard utilization typically shows an early peak then decays toward a plateau, and Soft utilization shows a mild overshoot before converging, yet late crossovers among the  widths still occur.

In summary, our analysis of spectral rank dynamics shows a consistent width power law that emerges after stabilization ($\sim$5K steps) with reliable fit quality ($R^2 \geq 0.6$). The stabilized exponents  for rank utilization ($\beta_{\text{hard}} < 0$ and $\beta_{\text{soft}} > 0$) highlight the key trade-off: increasing width reduces concentration in dominant modes while broadening soft spectral utilization. Although transient crossovers can appear in both raw and normalized ranks, they do not alter the exponent $\beta$ or the trade-off they encode. Thus, spectral scaling can be reliably characterized by the converged $\beta$ and $R^2$ values, providing a quantitative relation between FFN width and latent space utilization.

\section{Case Study for Spectral Rank}

\subsection{LayerNorm and Spectral Rank}

\begin{figure*} [t]
\centering
\subfloat[LLaMA-250M (PostLN) \label{subfig:llama_250m_postln_vanilla}]{\includegraphics[width=.34\textwidth]{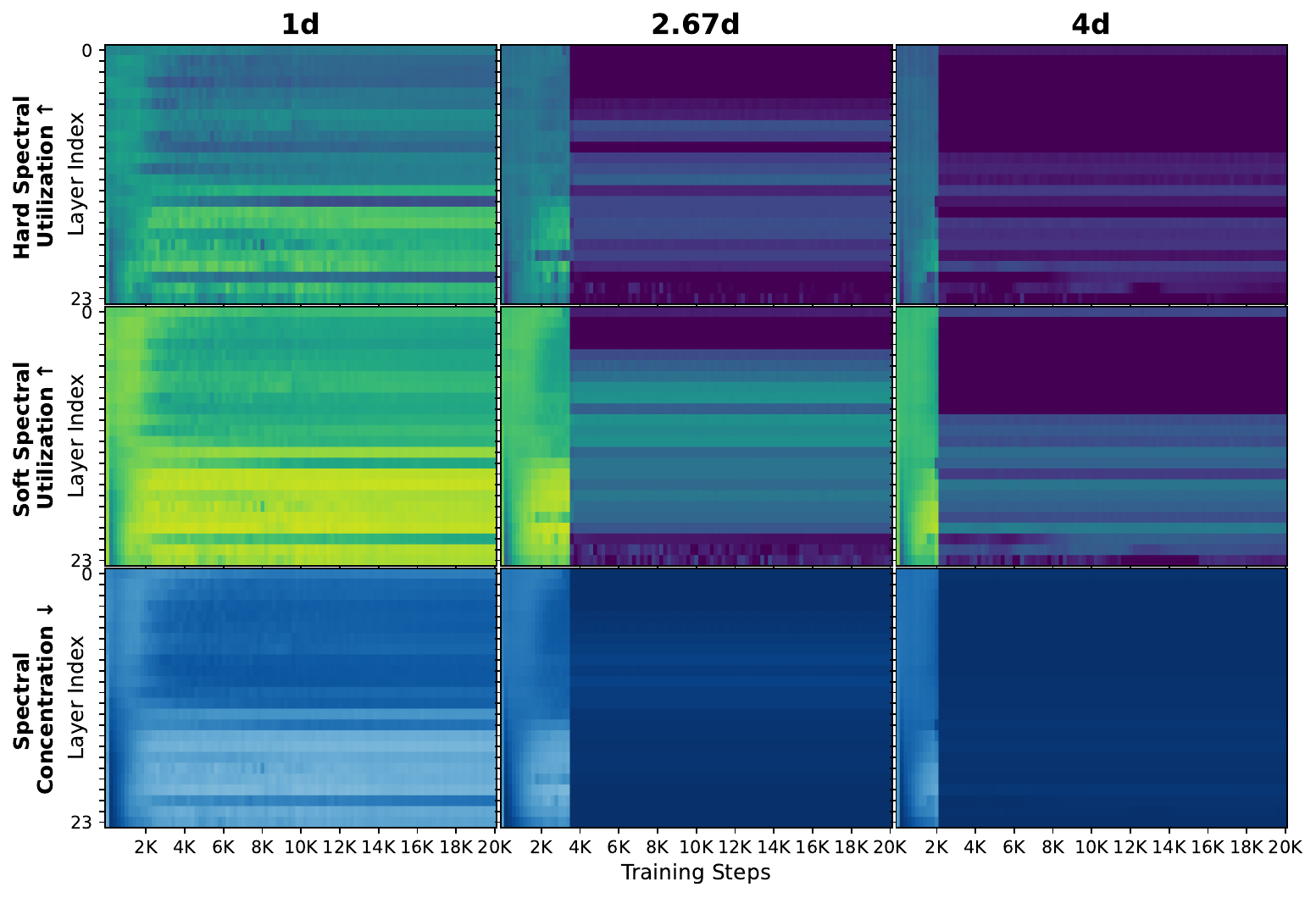}}
\subfloat[LLaMA-250M  (PostLN) +WNorm \label{subfig:llama_250m_postln_wnorm}]{\includegraphics[width=.32\textwidth]{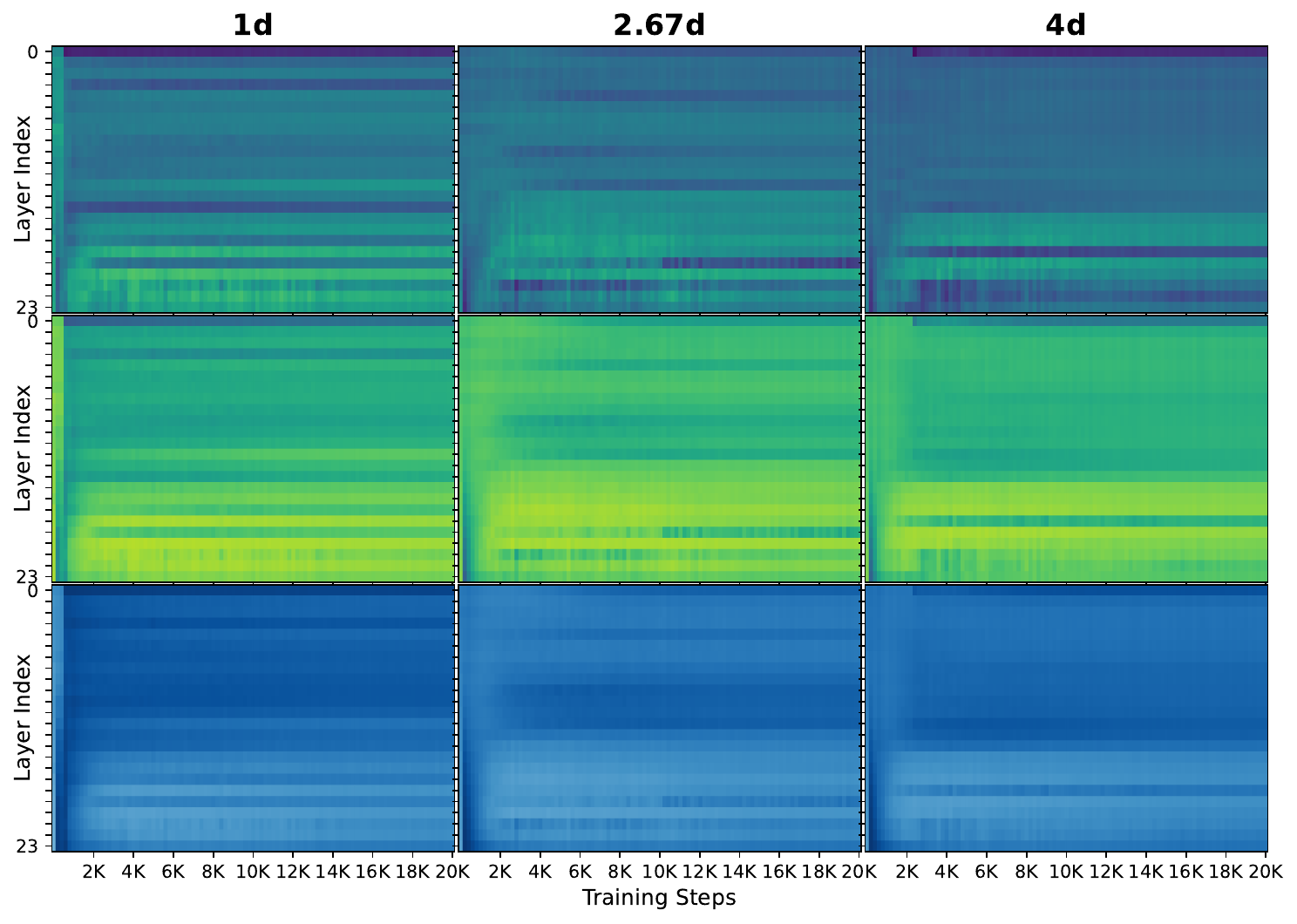}} 
\subfloat[LLaMA-250M (PostLN) + HNorm\label{subfig:llama_250m_postln_hnorm}]{\includegraphics[width=.335\textwidth]{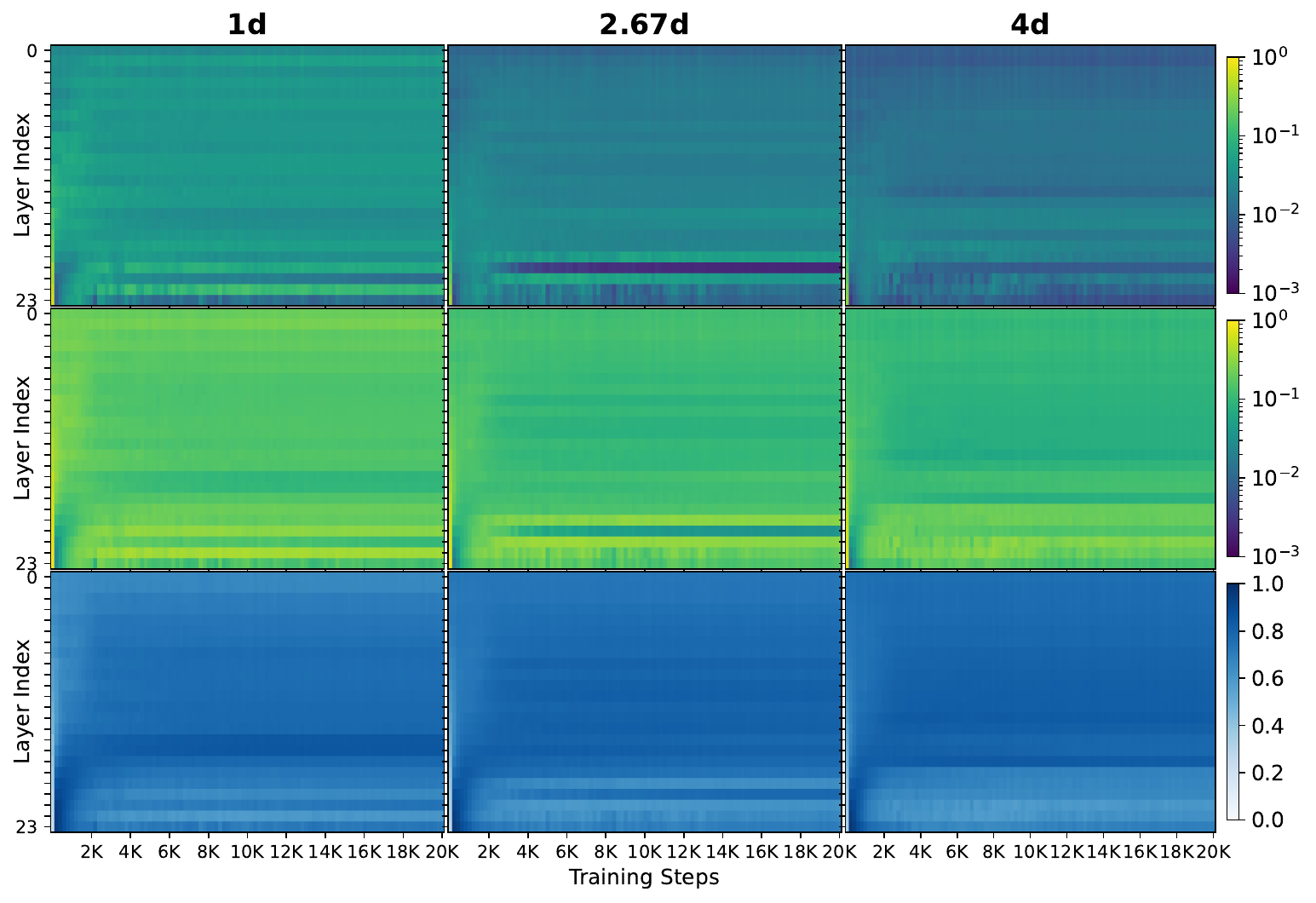}} 
\vspace{-0.5em}
\caption{Normalizing FFN weights stabilizes spectral dynamics in LLaMA-250M (PostLN).
Heatmaps show Hard Spectral Utilization (top), Soft Spectral Utilization (middle), and Spectral Concentration (bottom) across layers (y-axis) and training steps (x-axis) for $1d$, $2.67d$, and $4d$ FFN widths. Spectral utilization are shown in log scale while spectral concentration use linear scale. Vanilla PostLN becomes unstable and at higher widths, visible as darker regions for $2.67d$ and $4d$ in (a). Adding Weight Norm. (b) or Hyperspherical Norm. (c) to FFN linear layers stabilizes training, producing smoother spectral dynamics and more balanced hard- and tail-mode utilization.}
\label{fig:Panel_PreLLaMA250M}
\end{figure*}

\textbf{Pre-LN shows the classic asymmetry.}
With Pre-LN, soft-rank scales close to linearly with width ($\beta \approx 0.88$ at 70M; $\beta \approx 1.07$ at 130M, high $R^2$), while hard-rank is clearly sublinear ($\beta \approx 0.45/0.60$, lower $R^2$). This is the baseline \textit{tail-first growth}: widening expands low-energy directions, while the high-energy core lags behind (Table \ref{tab:spectral_scaling}).

\textbf{Post-LN suppresses tail growth.}
Shifting LayerNorm after the sub-blocks lowers soft-rank slopes to $\sim 0.71 - 0.82$ with stronger $R^2$, effectively \textit{dampening tail inflation}. Hard-rank slopes rise modestly to $\sim 0.52 - 0.56$ with better $R^2$, suggesting more orderly,but still sublinear,growth of the dominant subspace. Intuitively, normalizing after each transformation curbs variance spread, limiting activation of faint directions as width increases.

\textbf{Mix-LN balances core and tail.}
Mix-LN restores near-linear soft-rank scaling ($\beta \approx 0.97 - 1.10$, high $R^2$) while maintaining hard-rank growth above Pre-LN/Post-LN levels ($\beta \approx 0.59 - 0.63$, moderate $R^2$). In effect, it \textit{preserves tail coverage} while also improving dominant-mode scaling, avoiding both the over-tailing of Pre-LN and the excessive tail suppression of Post-LN.



\begin{table*}[htbp]
\centering
\caption{Spectral scaling law parameters ($\beta \pm$CI, $R^2$) for various LayerNorm positions (PreLN, PostLN, MixLN) across LLaMA models (70M, 130M, 250M). \textcolor{red}{Red boxes} highlight a significant improvement in MixLN hard rank scaling behavior. $^*$PostLN results for LLaMA-250M are unavailable due to training instability at higher FFN width.}
\label{tab:spectral_scaling}
\resizebox{0.99\textwidth}{!}{
\begin{tabular}{l*{6}{c}}
\toprule
& \multicolumn{2}{c}{PreLN} & \multicolumn{2}{c}{PostLN} & \multicolumn{2}{c}{MixLN} \\
\cmidrule(lr){2-3} \cmidrule(lr){4-5} \cmidrule(lr){6-7}
Model & Hard Rank & Soft Rank & Hard Rank & Soft Rank & Hard Rank & Soft Rank \\
\midrule
LLaMA-70M & \makecell{$0.451 \pm 0.778$ \\ $(R^2 = 0.251)$} & \makecell{$0.879 \pm 0.490$ \\ $(R^2 = 0.763)$} & \makecell{$0.556 \pm 0.358$ \\ $(R^2 = 0.706)$} & \makecell{$0.712 \pm 0.273$ \\ $(R^2 = 0.872)$} & \makecell{\fcolorbox{red}{white}{$0.593 \pm 0.668$} \\ \fcolorbox{red}{white}{$(R^2 = 0.440)$}} & \makecell{$0.972 \pm 0.477$ \\ $(R^2 = 0.805)$} \\[0.8em]
LLaMA-130M & \makecell{$0.604 \pm 0.411$ \\ $(R^2 = 0.684)$} & \makecell{$1.069 \pm 0.292$ \\ $(R^2 = 0.930)$} & \makecell{$0.521 \pm 0.294$ \\ $(R^2 = 0.758)$} & \makecell{$0.818 \pm 0.372$ \\ $(R^2 = 0.829)$} & \makecell{\fcolorbox{red}{white}{$0.626 \pm 0.484$} \\ \fcolorbox{red}{white}{$(R^2 = 0.626)$}} & \makecell{$1.096 \pm 0.484$ \\ $(R^2 = 0.837)$} \\[0.8em]
LLaMA-250M & \makecell{$0.407 \pm 0.671$ \\ $(R^2 = 0.268)$} & \makecell{$0.872 \pm 0.353$ \\ $(R^2 = 0.859)$} & \multicolumn{2}{c}{\makecell{$^*$\textit{Training Instability}}} & \makecell{\fcolorbox{red}{white}{$0.568 \pm 0.316$} \\ \fcolorbox{red}{white}{$(R^2 = 0.763)$}} & \makecell{$0.989 \pm 0.257$ \\ $(R^2 = 0.937)$} \\
\bottomrule
\end{tabular}}
\end{table*}


\subsection{LLaMA-250M PostLN}

{\em Spectral collapse in Post-LayerNorm blocks.}
We observe a strong correlation between spectral health and the  performance of LLaMA-250M when the FFN width is increased. In the vanilla Post-LayerNorm setup, spectral dynamics remain stable only for the narrowest FFN width (1d). However, scaling the width to 2.67d or 4d leads to a rapid collapse of spectral diversity: the hard-rank plunges to $\lesssim10^{-3}$ and the concentration saturates to $\approx1.0$ within the first few thousand steps (Figure \ref{subfig:llama_250m_postln_vanilla}). This spectral collapse signifies that most of the variance is funneled into one or two dominant directions, leaving the majority of the $\sim3000$ latent dimensions inactive. As a result, model performance deteriorates sharply, with test perplexity exceeding consistent with the figures reported in Table \ref{tab:llama250m_norm}.

\begin{table}[htbp]
\setlength{\tabcolsep}{4pt} 
\centering
    \caption{Vanilla PostLN in LLaMa-250M becomes unstable at higher FFN dimensions, causing spikes in PPL values. Adding Weight Normalization or Hyperspherical Normalization to the FFN linear layers stabilizes training (former outperforms the latter across all scales).}
    \label{tab:llama250m_norm}
    \begin{tabular}{l@{\hspace{3pt}}*{3}{S[table-format=4.2]}}
    \toprule
    PostLN & {$1d$} & {$2.67d$} & {$4d$} \\
    \midrule
    Vanilla      & 27.10  & 1427.91 & 1431.01 \\
    WeightNorm  & 28.89  & 25.08   & 24.27 \\
    HypersphericalNorm   & 31.66  & 27.92   & 26.48 \\
    \bottomrule
    \end{tabular}
\end{table}

{\em Weight Normalization enables high-rank spectra and best perplexity.}
Employing weight normalization (WNorm) \cite{salimans2016weight} within each FFN significantly mitigates this collapse. The hard-rank stabilizes in the $10^{-2}$–$10^{-1}$ range, while spectral concentration settles around 0.25–0.3, indicating that hundreds of latent directions carry meaningful variance. This richer and more distributed latent basis translates into notably better performance: perplexities of 25.1 (at 2.67d) and 24.3 (at 4d), both outperforming the vanilla 1d baseline (27.1). These results affirm that maintaining a non-degenerate spectrum not only prevents collapse but also improve model's performance.


\begin{figure*} [t]
\centering
\subfloat[ {\scriptsize GPT-2 Spectral Scaling}]{\includegraphics[width=.25\textwidth]{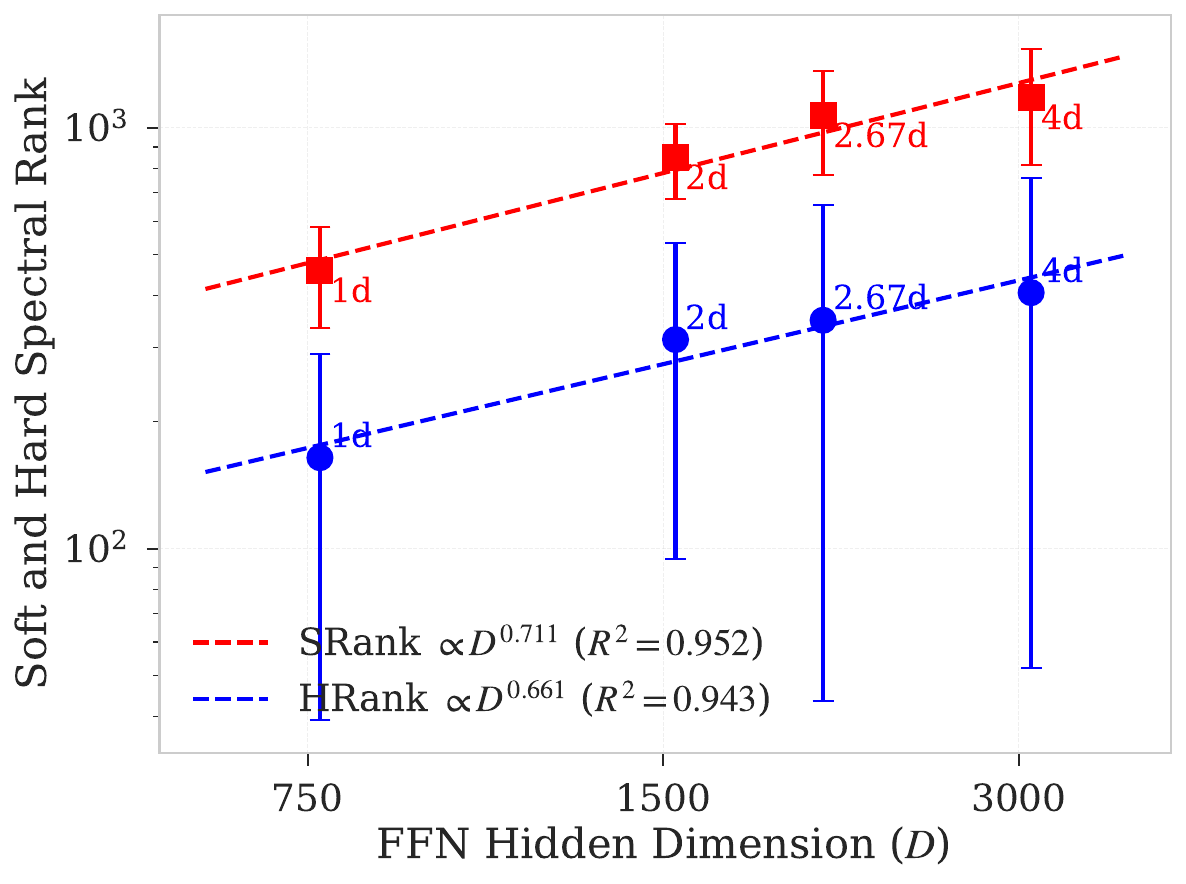}}
\subfloat[ {\scriptsize nGPT Spectral Scaling}]{\includegraphics[width=.25\textwidth]{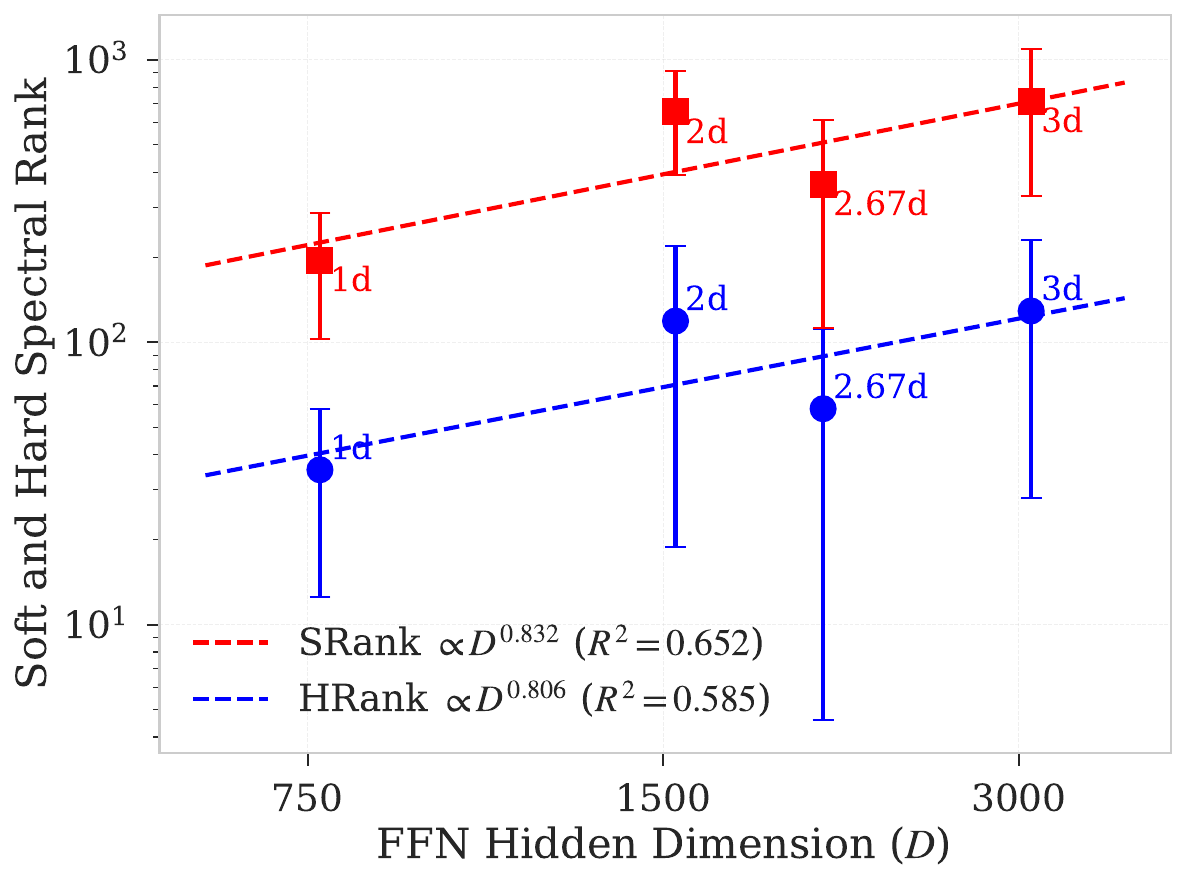}} 
\subfloat[{\scriptsize GPT-2 Spectral Utilization Scaling} ]{\includegraphics[width=.25\textwidth]{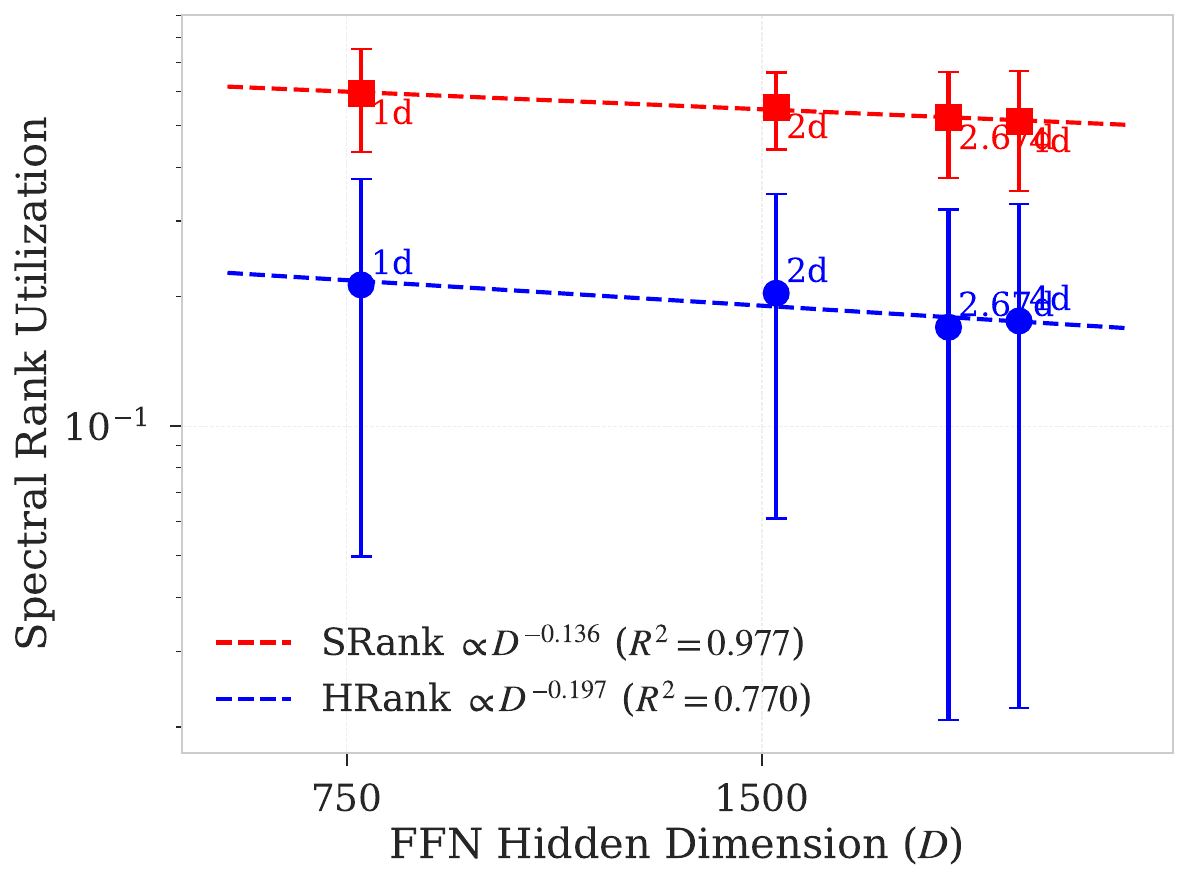}} 
\subfloat[ {\scriptsize nGPT Spectral Utilization Scaling} ]{\includegraphics[width=.25\textwidth]{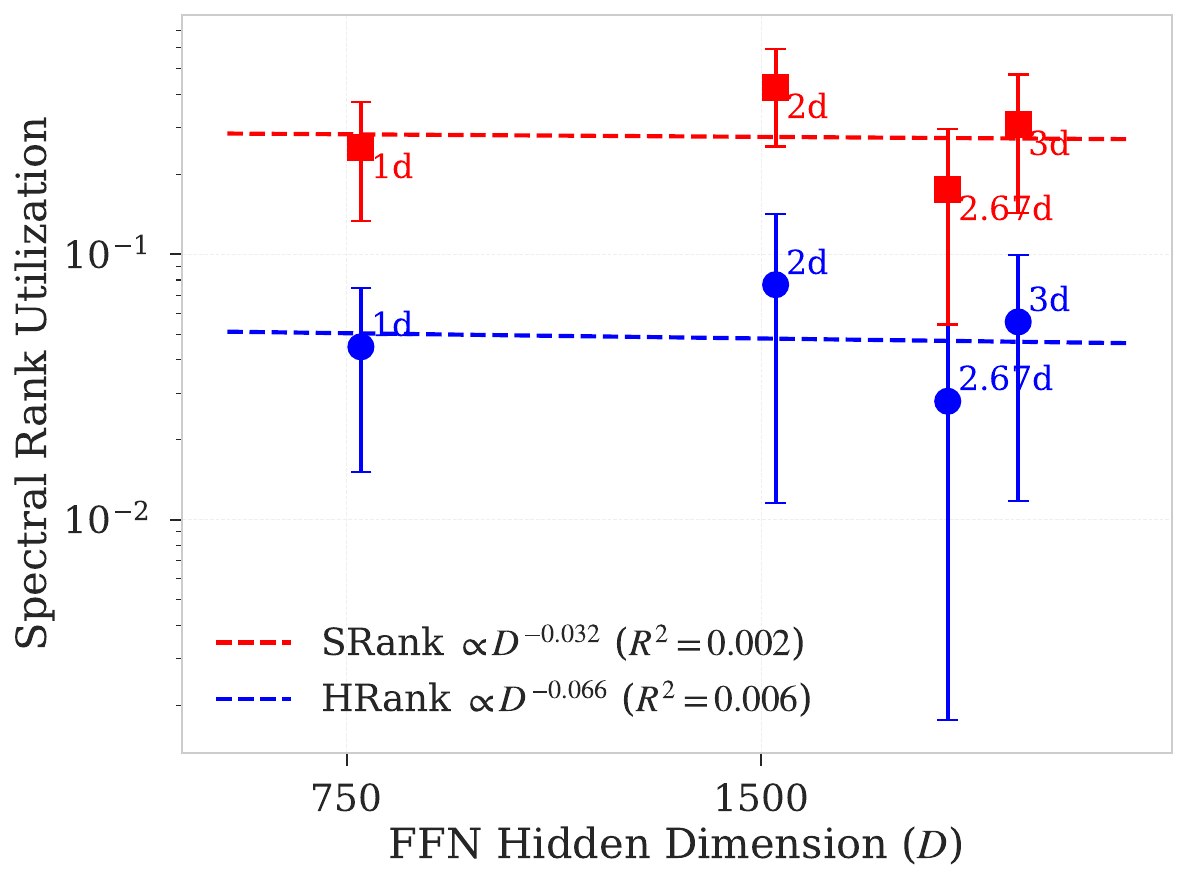}} 
\vspace{-0.5em}
\caption{Spectral rank and utilization vs. FFN width scaling in GPT-2 and nGPT. Panels (a,b) show raw ranks, while (c,d) plot normalized rank utilization for Soft rank (SRank, \textcolor{red}{\bf red}) and hard rank (HRank, \textcolor{blue}{\bf blue}) on log-log axes ($d$=768, width sweep $D \in \{1d, 2d, 2.67d, 3d\}$). {\em Hyperspherical constraints reduce soft-hard rank asymmetry, yielding balanced spectral dynamics and an effective utilization of FFN width  nGPT.}}
\label{fig:ScalingLaws_summary_gpt2_ngpt}
\end{figure*}



\begin{figure*} [t]
\centering
\subfloat[GPT-2 (GeLU) \label{subfig:gpt2_gelu}]{\includegraphics[width=.34\textwidth]{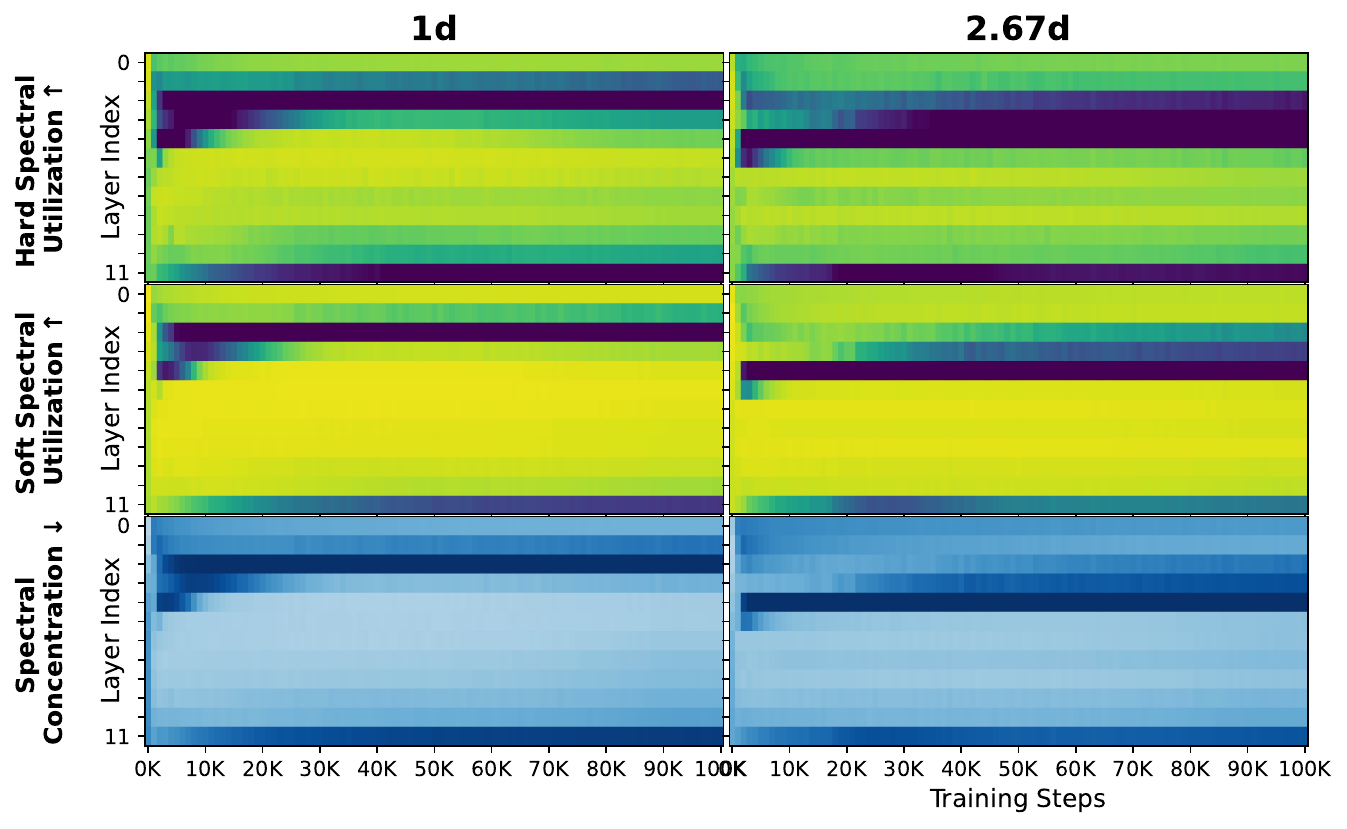}}
\subfloat[GPT-2 (SiLU) \label{subfig:gpt2_silu}]{\includegraphics[width=.32\textwidth]{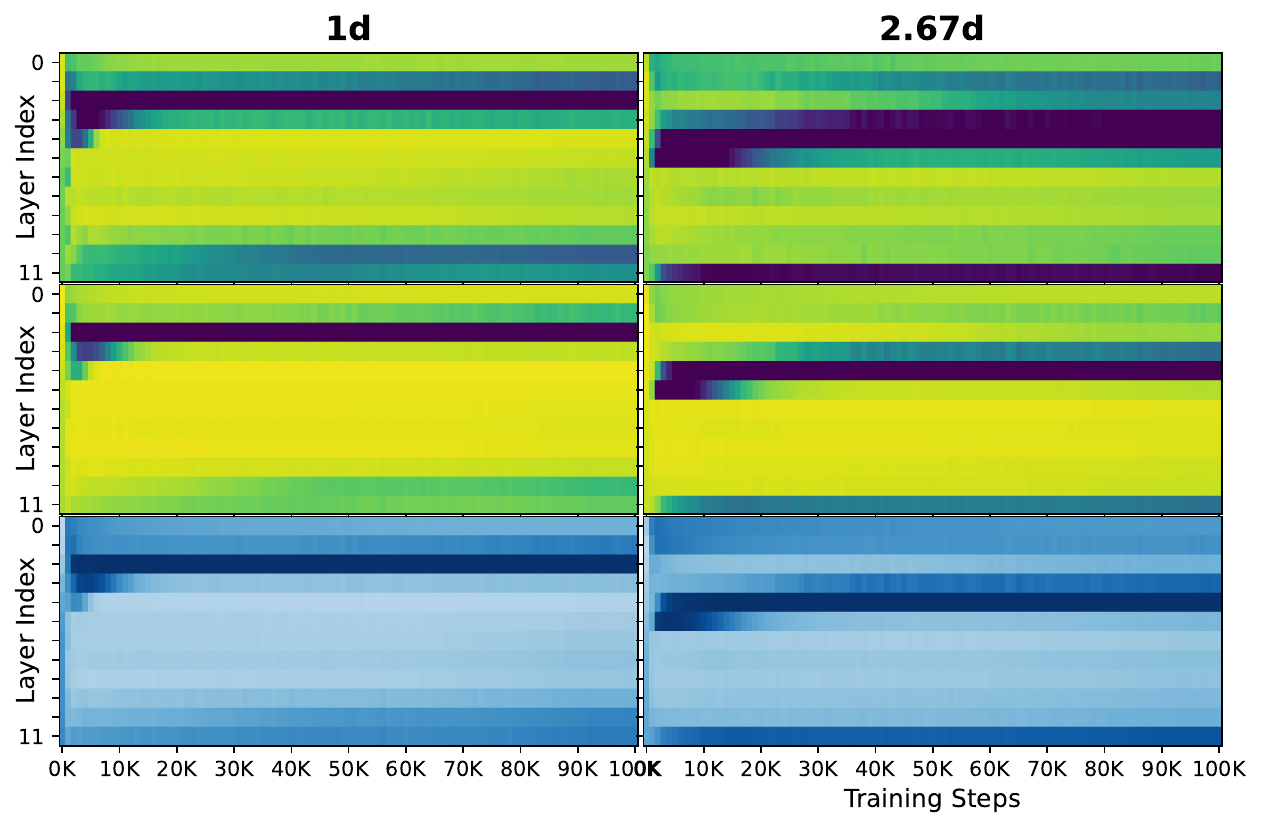}} 
\subfloat[nGPT-2 (SiLU)\label{subfig:ngpt}]{\includegraphics[width=.335\textwidth]{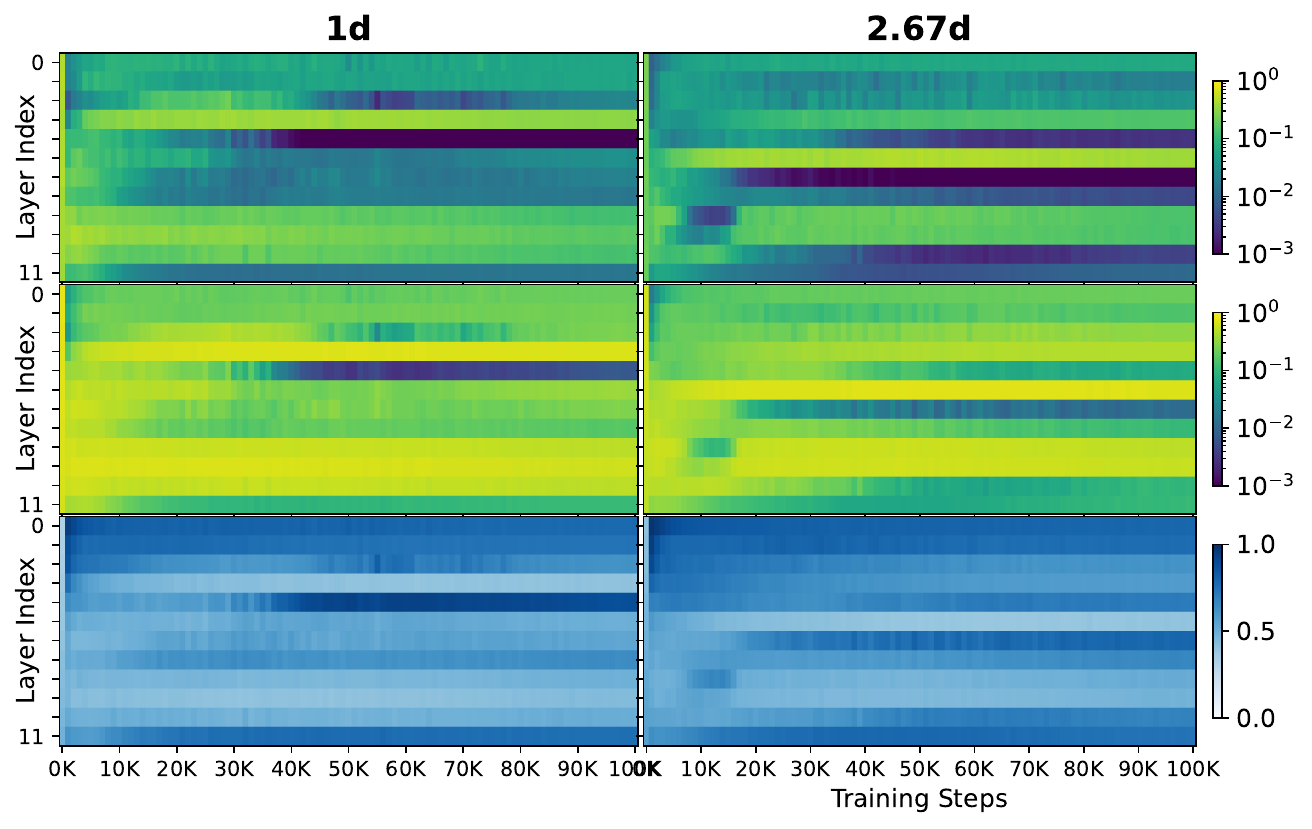}} 
\vspace{-0.5em}
\caption{Layer-wise spectral utilization dynamics (GPT-2 vs nGPT). Heatmaps show Hard Spectral Utilization (top), Soft Spectral Utilization (middle), and Spectral Concentration (bottom) across layers (y-axis) and training steps (x-axis). Each panel compares 1$d$ vs 2.67$d$ FFN width. Spectral utilization are shown in log scale (top two rows) while spectral concentration use linear scale. }
\label{fig:Panel_nGPT}
\end{figure*}

\subsection{Hyperspherical Normalization }

Hyperspherical normalization (HNorm) also prevents collapse and promotes training stability but results in more conservative spectral utilization \cite{loshchilov2025ngpt,lee2025hyperspherical,karras2024analyzing,wang2020understanding,liu2017deep}. The hard-rank remains roughly an order of magnitude above the collapse threshold, yet $\sim$30\% lower than the WNorm trace. Spectral concentration is marginally higher, suggesting a somewhat narrower effective basis. Consequently, while HNorm yields stable performance (27.9 at 2.67d and 26.5 at 4d), it does not match the perplexity gains achieved with WNorm. These findings highlight that collapse prevention is a necessary condition, but further lifting the rank and ensuring richer variance distribution is critical for unlocking full potential of wider FFNs.

{\bf Activation gating and normalization in GPT2.}
Figure \ref{fig:Panel_nGPT} tracks the spectral evolution, and Table \ref{tab:gpt2_ngpt_ppl} shows perplexity outcomes of GPT-2 variants using different activation and normalization schemes under two FFN widths (1d and 2.67d). The baseline GPT-2 with GeLU shows early hard-rank growth that quickly saturates around $10^{-2}$, while spectral concentration remains high ($\approx 0.7$). This indicates a narrow set of dominant directions and leads to moderate perplexity (14.07 at 2.67d), with limited gain over the 1d baseline (15.63).


\begin{table} [htbp]
\centering
\caption{Perplexity (PPL) comparison of GPT-2 and nGPT \cite{loshchilov2025ngpt} with different activation functions and FFN dimensions.}
\resizebox{0.49\textwidth}{!}{
\begin{tabular}{l *{6}{c}}
\toprule
& \multicolumn{2}{c}{GPT-2(GeGLU)} & \multicolumn{2}{c}{GPT-2(SwiGLU)} & \multicolumn{2}{c}{nGPT(SwiGLU)} \\
\cmidrule(lr){2-3} \cmidrule(lr){4-5} \cmidrule(lr){6-7}
 & 1d & 2.67d & 1d & 2.67d & 1d & 2.67d \\ \toprule
PPL & 15.63 & 14.07 & 15.60 & 14.05 & 15.01 & 13.60 \\ \bottomrule
\end{tabular}}
\label{tab:gpt2_ngpt_ppl}
\end{table}


The nGPT configuration augments SwiGLU with hyperspherical weight and activation normalization and a learnable residual eigen-learning rate  \cite{loshchilov2025ngpt}. This combination substantially enhances spectral performance: hard-rank remains two orders of magnitude above collapse, soft-rank saturates earlier with less fluctuation, and concentration reduces to $\approx 0.4$, a 20\% improvement over GPT-2. These gains are mirrored in performance, with perplexity dropping to 13.60 at 2.67d and stabilizing to 15.01 at 1d, outperforming both prior setups.

{\bf Hyperspherical learning reduces asymmetry and converts width into shared capacity.}
Across the width sweep, vanilla GPT-2 shows the familiar split: hard rank (dominant modes) saturates early, while soft rank (tail) keeps rising, so added dimensions drift into the tail. With hyperspherical constraints (nGPT), the soft-hard gap narrows in both raw ranks and normalized utilization: slopes move closer and the separation between the soft and hard curves decreases (Figure \ref{fig:ScalingLaws_summary_gpt2_ngpt}).

In practice, nGPT sustains growth in dominant modes instead of stalling, while tails expand without overwhelming the spectrum, yielding a more balanced distribution. Moreover, their normalized utilization shows that GPT-2 dynamics remain uneven, whereas in nGPT they flatten into near-straight lines, indicating that FFN width is actually being used rather than pooled in the tail. This makes hyperspherical learning \citep{liu2021learning,liu2018learning,liu2017deep,wang2020understanding,lin2020regularizing,bernstein2025manifolds} a {\em promising representational technique for improving FFN latent-space utilization}, enabling more balanced spectral dynamics and efficient use of width.


\section{Conclusion}

We reframed FFN width selection as a spectral utilization problem, showing that widening follows a consistent tail-first pattern: soft-rank utilization remains near-linear while hard-rank utilization declines. This asymmetry, formalized as spectral scaling laws, reveals two efficiency failures, spectral dilution and spectral collapse, that limit naïve width growth. LayerNorm placement modulates these dynamics: Pre-LN amplifies tails, Post-LN suppresses them, and Mix-LN balances both. Together, these results highlight spectral utilization as a new efficiency axis, motivating width-efficient designs via layer-wise scheduling and pruning.

\section*{Limitations}
The study is limited to English decoder-only models up to 250M parameters and does not validate spectral behavior in multilingual or encoder-decoder settings. While spectral metrics correlate with perplexity, causality remains unproven, and finer-grained subspace analysis may be needed beyond scalar metrics like SUI. Additionally, eigen-computations could pose challenges at extreme scales.



\bibliography{custom}

\newpage
\appendix


\begin{table*}[htbp]
\setlength{\tabcolsep}{3.5pt} 
\centering
    \caption{Evaluation perplexity (PPL) for LLaMA models across different normalization positioning and FFN dimensions. The columns $1d$, $2.67d$, $4d$, and $6d$ represent different FFN width, where $d$ is the model dimension. The unusually high PPL in PostLN LLaMA-250M indicate training instability.}
    \label{tab:EvalPPlLLaMA}
    \resizebox{0.99\textwidth}{!}{
    \begin{tabular}{l@{\hspace{1pt}}*{12}{S[table-format=4.1]}} 
    \toprule
    \multirow{2}{*}{\textbf{Model}} & \multicolumn{4}{c}{\textbf{PreLN}} & \multicolumn{4}{c}{\textbf{PostLN}} & \multicolumn{4}{c}{\textbf{MixLN}} \\
    \cmidrule(lr){2-5} \cmidrule(lr){6-9} \cmidrule(lr){10-13}
    & {$1d$} & {$2.67d$} & {$4d$} & {$6d$} & {$1d$} & {$2.67d$} & {$4d$} & {$6d$} & {$1d$} & {$2.67d$} & {$4d$} & {$6d$} \\ 
    \midrule
    \textsc{LLaMA-70M}  & 38.6 & 34.2 & 32.4 & 31.1 & 38.2 & 33.6 & 32.3 & 31.1 & 38.7 & 33.9 & 32.0 & 30.7 \\
    \textsc{LLaMA-130M} & 29.6 & 26.4 & 25.8 & 24.6 & 29.2 & 26.7 & 25.8 & 25.1 & 29.2 & 26.8 & 25.3 & 24.3 \\
    \textsc{LLaMA-250M} & 26.7 & 24.5 & 23.3 & 22.5 & 27.1 & \bfseries{1427.9} & \bfseries{1431.0} & \bfseries{1436.7} & 26.8 & 24.2 & 23.0 & 22.5 \\     \bottomrule
    \end{tabular}}
\end{table*}


\begin{figure*} [t]
\centering
\subfloat[LLaMA-70M (PreLN) \label{subfig:llama_70m_pre}]{\includegraphics[width=.5\textwidth]{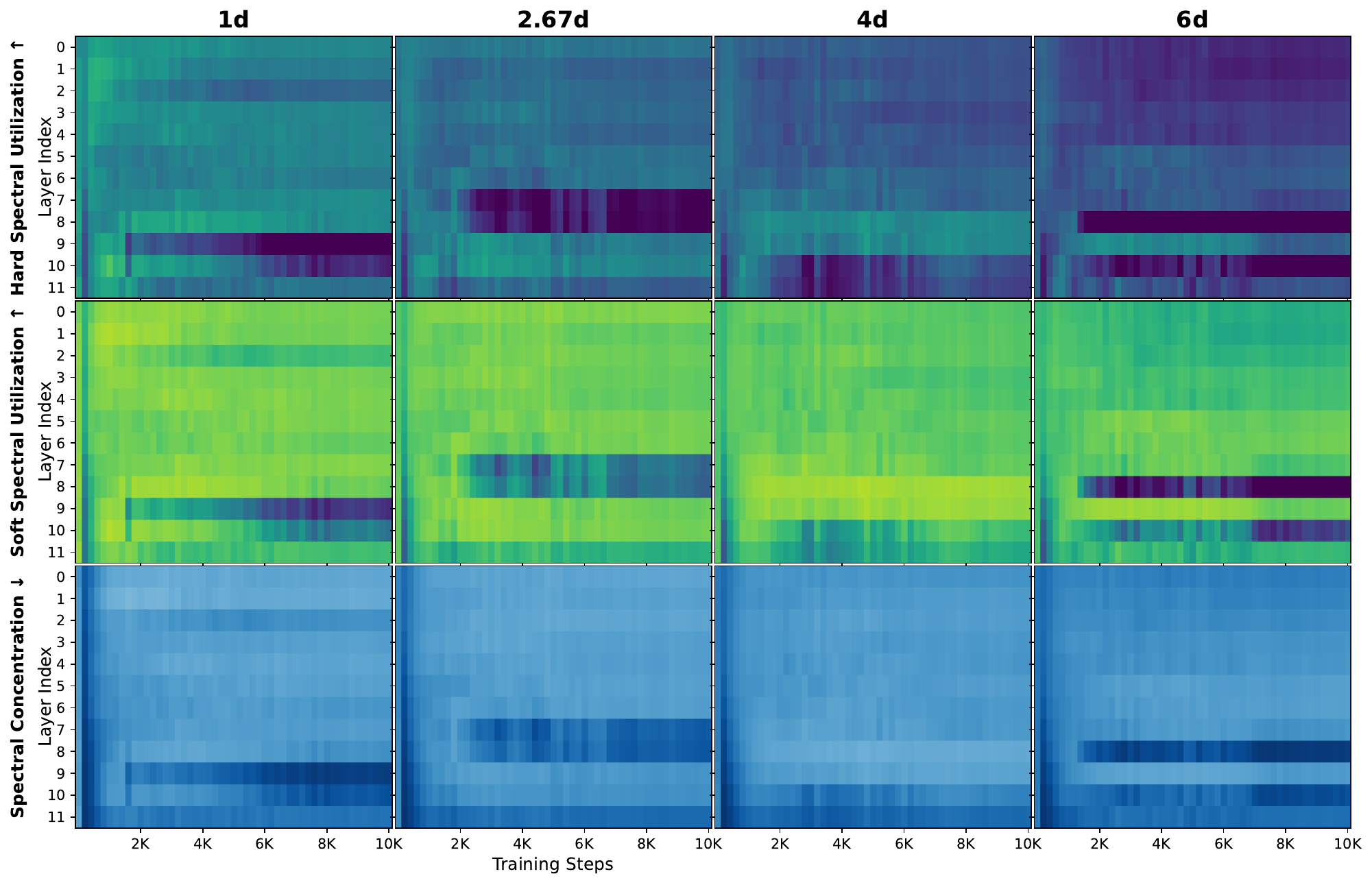}}
\subfloat[LLaMA-130M (PreLN) \label{subfig:llama_130m_pre}]{\includegraphics[width=.5\textwidth]{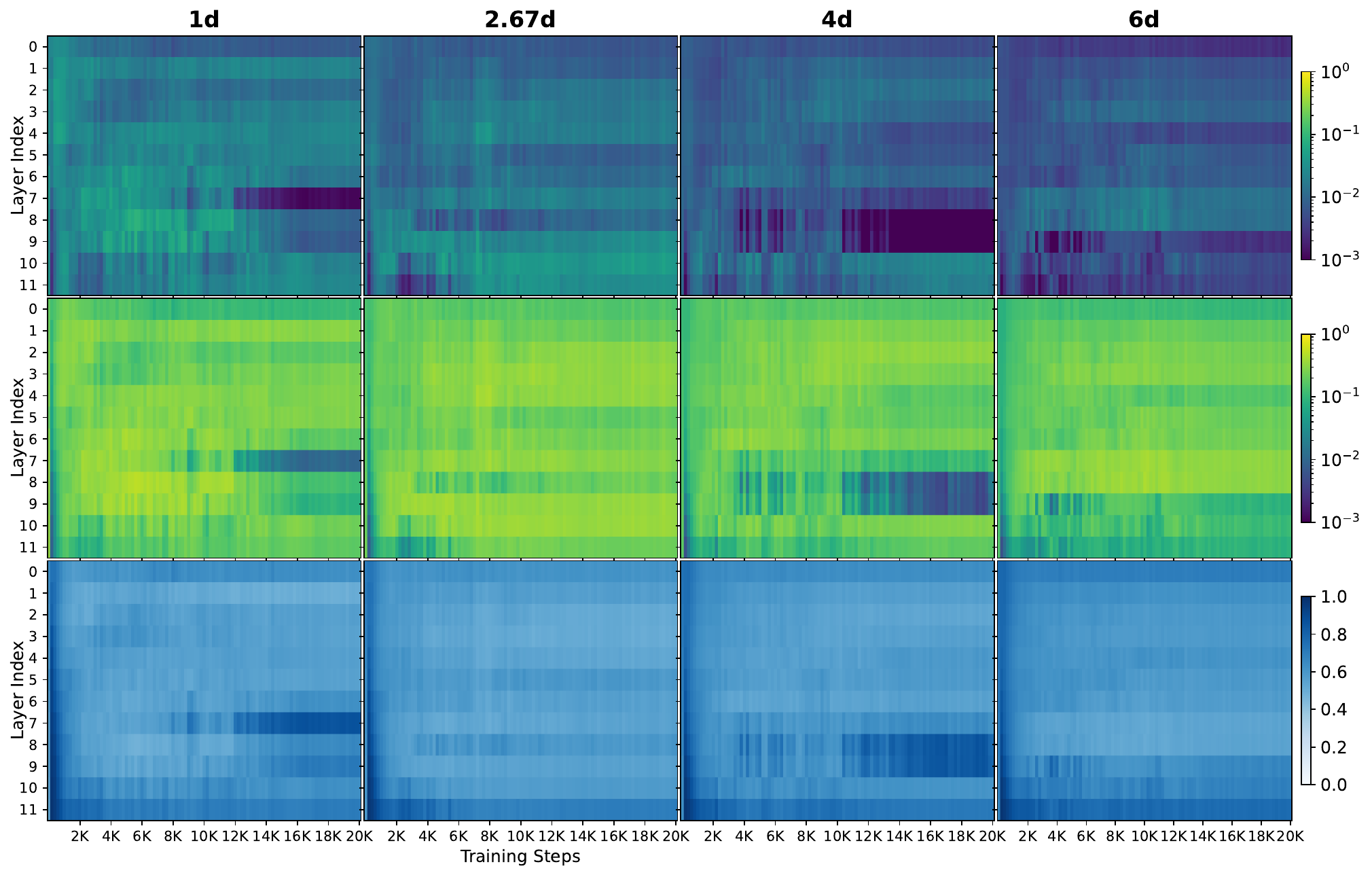}} \\ \vspace{-1em}
\subfloat[LLaMA-250M (PreLN) \label{subfig:llama_150m_pre}]{\includegraphics[width=.99\textwidth]{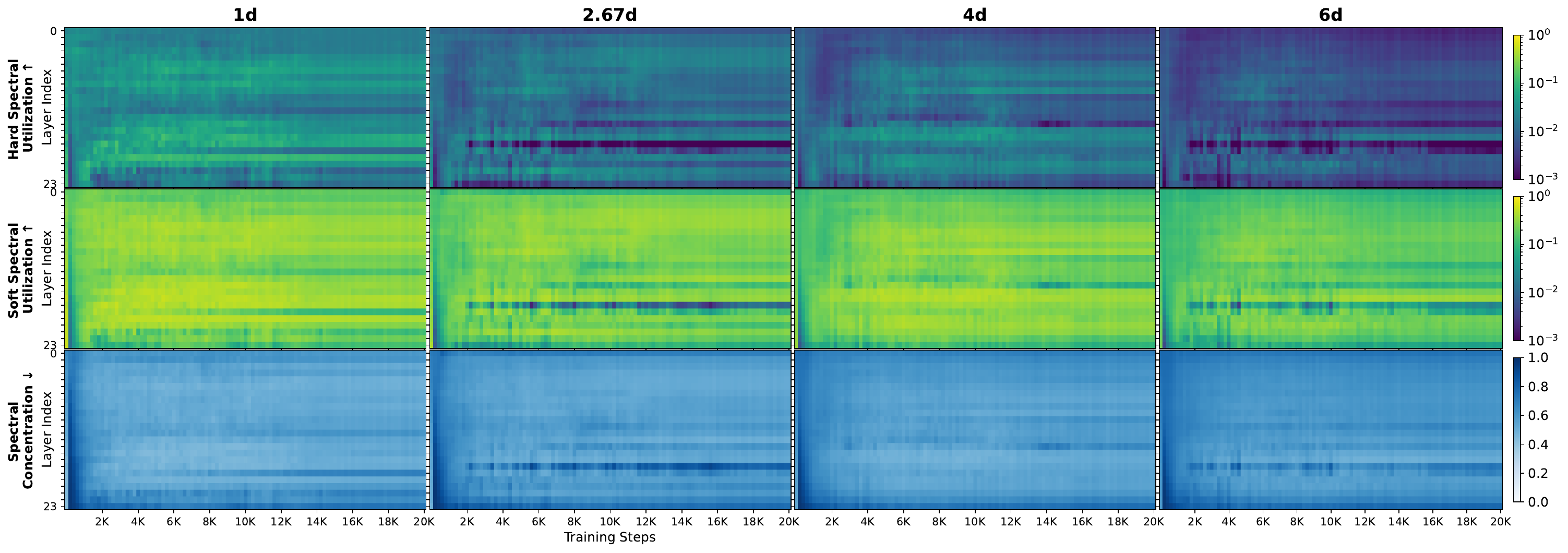}} 
\vspace{-0.5em}
\caption{LLaMA models}
\label{fig:panel_eigenmetrics_all_llama}
\end{figure*}


\end{document}